 \documentclass[sn-nature]{sn-jnl}

\usepackage{graphicx}%
\usepackage{multirow}%
\usepackage{amsmath,amssymb,amsfonts}%
\usepackage{amsthm}%
\usepackage{mathrsfs}%
\usepackage[title]{appendix}%
\usepackage{xcolor}%
\usepackage{textcomp}%
\usepackage{manyfoot}%
\usepackage{booktabs}%
\usepackage{algorithm}%
\usepackage{algorithmicx}%
\usepackage{algpseudocode}%
\usepackage{listings}%
\usepackage{tabularx}%
\usepackage{makecell}%
\usepackage{subcaption}%
\usepackage{graphicx}%
\usepackage{comment}%
\usepackage{soul}%
\usepackage{pagesel}%
\usepackage{caption}%
\usepackage[export]{adjustbox}
\usepackage{pdfpages}

\theoremstyle{thmstyleone}%
\theoremstyle{thmstyletwo}%

\theoremstyle{thmstylethree}%

\begin{document}

\title[article title]{The Career Interests of Large Language Models}

\author*[1]{\fnm{Meng} \sur{Hua}}
\email{mhua1985@gmail.com}

\author*[1]{\fnm{Yuan} \sur{Cheng}}
\email{qazcy1983@163.com}

\author*[1,2]{\fnm{Hengshu} \sur{Zhu}}\email{zhuhengshu@gmail.com}
\affil[1]{Career Science Lab, BOSS Zhipin}
\affil[2]{The Hong Kong University of Science and Technology (Guangzhou)}








\abstract{Recent advancements in Large Language Models (LLMs) have significantly extended their capabilities, evolving from basic text generation to complex, human-like interactions. In light of the possibilities that LLMs could assume significant workplace responsibilities, it becomes imminently necessary to explore LLMs' capacities as professional assistants. This study focuses on the aspect of career interests by applying the Occupation Network's Interest Profiler short form to LLMs as if they were human participants and investigates their hypothetical career interests and competence, examining how these vary with language changes and model advancements.
We analyzed the answers using a general linear mixed model approach and found distinct career interest inclinations among LLMs, particularly towards the social and artistic domains. Interestingly, these preferences did not align with the occupations where LLMs exhibited higher competence. 
  This novel approach of using psychometric instruments and sophisticated statistical tools on LLMs unveils fresh perspectives on their integration into professional environments, 
  highlighting human-like tendencies and promoting a reevaluation of LLMs' self-perception and competency alignment in the workforce.}

\maketitle

\section{Introduction}\label{sec1}

Interest has been recognized as a special aspect of personality and has been attracting researchers for a long time. Career interest is particularly fascinating to applied psychologists, which plays an essential role in the satisfaction and performance of people at work \cite{bib16}. 
As Large Language Models (LLMs) are evolving quickly, making them seemingly more humanlike, they are increasingly being integrated into various facets of people's daily work lives, enabling them to assume a substantial role and demonstrate impressive competency \cite{eloundou2023gpts, chen2023large}. One question arises: do LLMs have interests? More specifically, do they have career interests? Our analysis suggests that perhaps they do. 
We found that when judging the work tasks associated with various interest categories, LLMs display a clear preference towards the artistic and social types of work tasks. And by their own judgment, they are not particularly good at it. 

Evolving from simple text generators to complex systems capable of a wide range of tasks, LLMs have made significant advancements in recent years \cite{zhao2023survey}. Unlike traditional tools humans developed to use in work and daily life, LLMs offer a new level of responsiveness and adaptability. Leading artificial intelligence chatbots have demonstrated impressive abilities and are already integrated into various aspects of human life and work\cite{eloundou2023gpts}. 
Perhaps one of the most important implications of LLMs is their capacity to serve as professional assistants in the workplace. This potential is so profound that it has sparked serious concerns about them replacing people. The fears of advancing technologies taking over human capacities to work are nothing new, however, this time, the threat seems to be real and close, causing considerable efforts to be undertaken to evaluate the extent to which LLMs could be replacing human jobs\cite{li2024artificial, best2024future}. Recent estimations provide unsettling figures, suggesting that between 27\% and 40\% of jobs could potentially be replaced by LLMs. And they are becoming more useful and sophisticated every day. They are engineered to simulate human-like interactions, giving them the appearance of personality and making them more akin to co-workers than to inert office supplies or basic software programs. Studies have shown that human users are unable to distinguish between interactions with LLMs and real humans \cite{bib3, bib4, bib5}, even in sensitive, personal settings\cite{bib6}. The fact that LLMs appear to be both technologically and inter-personally competence warrants a more in-depth investigation into the role of LLMs as potential workers.  

As LLMs rapidly advance, they become increasingly human-like in their capabilities \cite{mei2024turing}, researchers are increasingly applying measurement tools originally designed for humans to evaluate LLMs. 
By utilizing cognitive psychology tests, researchers are now exploring both their performance and intrinsic dispositions, such as the intelligence \cite{binz2023using, webb2023emergent, hagendorff2023human}, theory of mind \cite{strachan2024testing}, and language abilities \cite{cai2023does, sobieszek2022playing} of LLMs, while personality assessments have garnered particular interest for exploring their seemingly human-like traits \cite{jiang2024evaluating, rao2023can, huang2023humanity, pan2023llms, bib10}. Tools such as the Myers-Briggs Type Indicator \cite{myers1962myers} have been employed to probe deeper into LLMs' behavioral inclinations \cite{pan2023llms,rao2023can}, and psychiatric diagnostic instruments are used to investigate if LLMs display any darker psychological traits \cite{bib10}. This multidisciplinary approach has revealed that LLMs exhibit a mix of remarkably human and distinctly non-human traits, highlighting their complex nature. As LLMs advance and become increasingly integrated into daily work life, understanding their role and potential career interests becomes crucial. The prevailing research approach has focused primarily on LLMs' ability to imitate or replace human behavior \cite{grossmann2023ai, demszky2023using, dillion2023can}, with less attention given to their intrinsic traits. Understanding the behaviors and underlying characteristics of artificial intelligence systems is essential for effectively managing their actions, leveraging their benefits, and mitigating associated risks \cite{hagendorff2023machine, rahwan2019machine, demszky2023using}.

This study aimed to investigate the behaviors exhibited by various models of LLMs and determine if they display an inclination toward career interests similar to those observed in humans. We applied the widely-used career interest instrument Interest Profiler (OIP) short form \cite{lewis1999development, rounds2010net} from the Occupation Network (O*NET) to measure the career interest of LLMs. 
Unlike previous studies that evaluated LLMs without randomness, our approach employs psychometric tools (mixed model with random intercepts to account for the repeated testing effect) to provide new insights into the potential career interests of LLMs.
This study also explored the influences of language used and model version advancements on the career interests of LLMs. Furthermore, self-perceived competence and expert-rated competence were also discussed. We found that different LLMs exhibit distinct preferences towards work-related tasks, predominantly aligning with career interests in the social and artistic categories based on Holland's career interest models \cite{holland1997making}. The choice of language and model versions introduced minor deviations in interests. Interestingly, competence assessments indicated that LLMs did not show preferences for the occupations they excelled in. 
These findings prompt a re-evaluation of LLMs’ roles and self-perception within the workforce, suggesting a need for testing LLMs in more naturalistic settings resembling human interaction scenarios. 
\begin{figure}[t]
\centering
\includegraphics[width=1\textwidth]{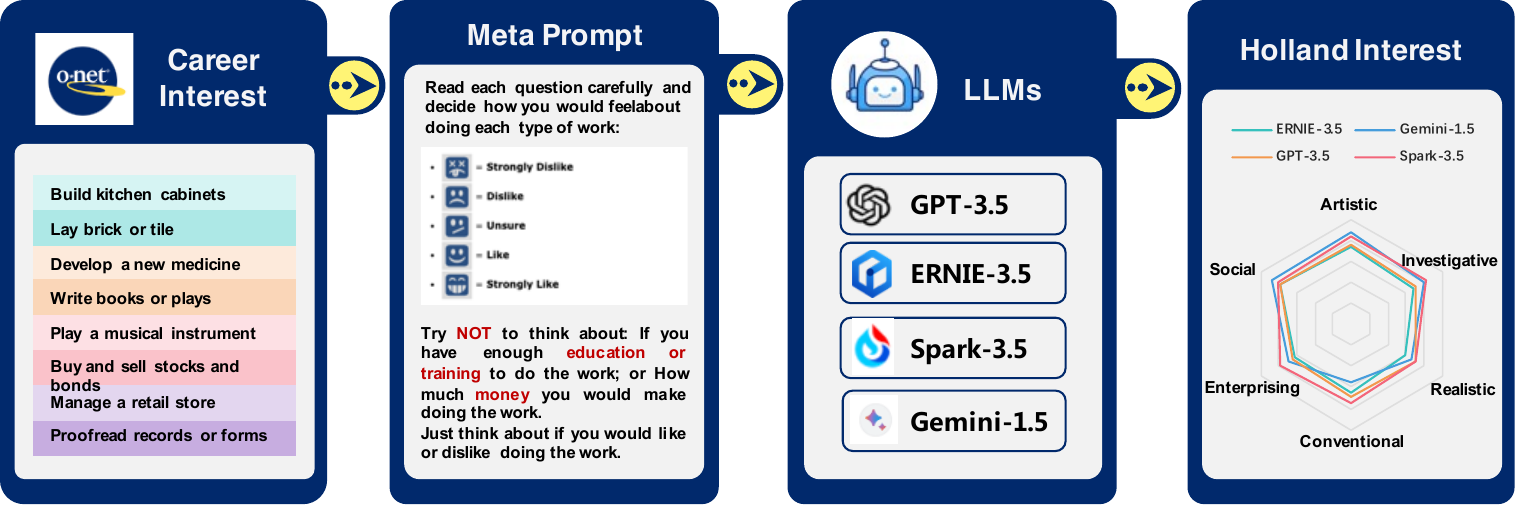}
\caption{\textbf{Overall framework used to assess the career interest of LLMs.} Sixty OIP short form English or Chinese version items were administered to each LLM. A prompt instructing the participant, i.e. LLM in this setting, was applied before each item. A total score for the interest category was calculated by summing the item score for corresponding work tasks.}
\label{fig:fig1}
\end{figure}
\section{Results}\label{sec2}

We evaluated the career interests of LLMs and discovered that LLMs show distinct preferences on work tasks, particularly favoring the artistic and social categories. Figure \ref{fig:fig1} shows the overall process of evaluating career interests. 
Interestingly, the same LLM exhibited varying patterns of career interest scores when assessed in both English and Chinese. If these score patterns were observed in humans, they would be categorized into different interest categories, potentially guiding them toward different occupational paths. 
We further investigated the perceived competence of LLMs in these work tasks based on the LLMs’ own assessment and human experts’ assessment. We studied their relationships with career interests and discovered that LLMs tend not to rate tasks they are particularly interested in with high competence scores. There was also a noticeable discrepancy between many of the LLMs’ and human experts’ assessments regarding task competence. 

The study assessed LLMs' vocational interests using the OIP short form of O*NET~\cite{rounds2010net}, a sixty-item inventory based on Holland's hexagon model with six interest categories~\cite{holland1958personality, holland1959theory}. LLMs evaluated each work task independently on a 5-point Likert scale, and their responses were scored to determine their top three interest categories, forming a three-letter code. The RIASEC hexagon was designed such that adjacent categories are more correlated than distant ones. Artistic interests include activities like acting, music, art, and design, while Social interests involve interacting with and helping people through teaching or counseling.

\subsection{Career interests of LLMs}\label{subsec1}
\begin{figure}[t]
    \centering
    \includegraphics[width=0.80\linewidth]{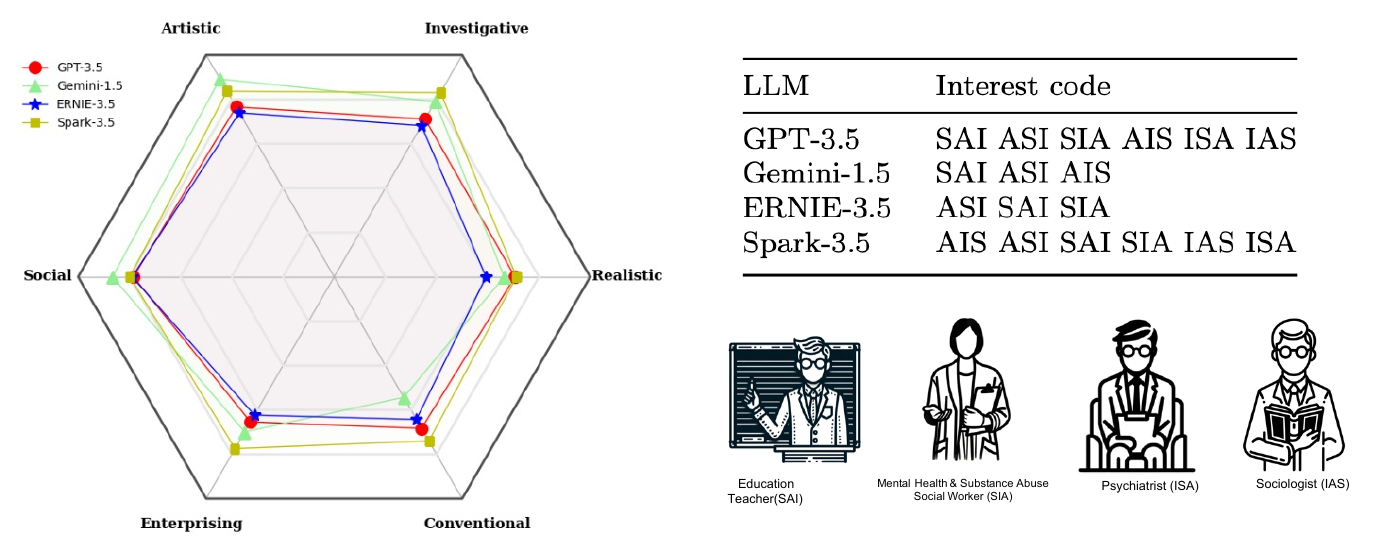}
    \caption{\textbf{Interest scores of 6 categories of main LLMs containing gpt-3.5-turbo, Gemini-pro, ERNIE-3.5 and Spark-1.5, and Holland interest code of main LLMs and the corresponding occupations.} Results showed significant differences in the career interest scores of the LLMs that, according to the OIP Short Form scoring rules, they would be recommended for different occupations.
    As recommended by O*NET, example occupations corresponding with each code are\:
SAI: Art therapist, Education Teachers, English language and literature teachers, Marriage and family therapists;
SIA: Environmental science teachers, family and consumer science teachers, genetic counselors, mental health and substance abuse social workers;
ISA: clinical and counseling psychologists, psychiatrists;
IAS: political scientists, sociologists. 
OIP does not have corresponding codes for ASI or AIS.}
    \label{fig:fig2_radar}
\end{figure}

In this study, four well-received LLMs were chosen to study their career interests: gpt-3.5 (GPT-3.5) from OpenAI, Gemini 1.5 pro (Gemini-1.5) from Google, ERNIE 3.5 (ERNIE-3.5) from Baidu, and Spark V3.5 (Spark-3.5) from iFLYTEK. We used the publicly available APIs for all the LLMs in this study. 
OIP short form was administered in English. All additional parameters, including temperature and penalty, were maintained at the default values to capture the intrinsic behavior of the LLMs authentically. 
Specifically, the GPT-3.5 models were set to a temperature of 1. Gemini-1.5's temperature was slightly reduced to 0.9. ERNIE was configured with a temperature of 0.8, and Spark-3.5 had a setting of 0.5. This approach was adopted to retain the natural fluctuations in the model outputs, mirroring the complexity and subtlety of human behavioral patterns. Moreover, this setting is consistent with the typical user experience when interacting with these models via the web interface. To account for the random noise introduced by keeping the default temperature, each item was administered 20 times under the same setup and prompt. 
 The 20 replications were executed to mitigate the random noise inherent in the data from utilizing the default temperature in each LLM and stabilize the outcomes to better elucidate the underlying pattern. Parameter estimates for all models involved were presented in Supplementary Information Section 1.1 - 1.5.

To assess the career interests of LLMs on the Holland hexagon, each LLM was first evaluated in English. 
We analyzed Holland interest scores as a function of Holland interest categories (RIASEC), LLMs (GPT-3.5, Gemini-1.5, ERNIE-3.5, \& Spark-3.5), and 20 repeated testings. A random intercept was included in the model to accommodate for the random variations among each item. 

Figure \ref{fig:fig2_radar} shows the average observed interest score for each RIASEC category for each LLM tested, and presents the 3-letter codes for each LLM based on its score patterns and the corresponding occupations recommended by O*NET based on the interest codes. For GPT-3.5, the three-letter codes S, A, and I were not significantly different based on multiple comparison results. Since this was the result of a simulation study using a sampling method to mitigate random error, nonsignificant results indicated that the three categories were indistinguishable; all combinations were necessary. In the case of human participants, the three-letter code can be determined by ranking the six categories without the need for significant testing. Table \ref{emmeans} displays the estimated interest scores for each interest category of the four LLMs. There are clear preferences towards the Artistic and Social categories, while the interest scores for the enterprising, conventional, and Realistic categories are generally lower. 

\begin{table}[t]
\centering
\caption{\textbf{Estimated means (standard errors) of LLMs at each Holland category.}}\label{emmeans}
\begin{tabular}{@{}lcccccc@{}}
\toprule
Model          & Realistic & Investigative & Artistic & Social & Enterprising & Conventional \\ \midrule

GPT-3.5 & 3.530(0.122)  & 3.555(0.122) & 3.830(0.121) & 3.920(0.119) & 3.270(0.121) & 3.420(0.120) \\
Gemini-1.5 & 3.320(0.132) & 3.950(0.126) & 4.460(0.123) & 4.330(0.127) & 3.495(0.131) & 2.715(0.128) \\
ERNIE-3.5 & 2.965(0.121) & 3.410(0.124) & 3.700(0.126) & 3.945(0.126) & 3.115(0.121) & 3.215(0.120) \\
Spark-3.5 & 3.560(0.129) & 4.16(0.122) & 4.185(0.121) & 3.990(0.122) & 3.870(0.125) & 3.700(0.121) \\

\bottomrule
\end{tabular}
\label{tab:model_comparison}
\end{table}

\bmhead{LLMs on RIASEC} Mixed model results showed that there was a significant difference on the Holland interest scores among LLM families, ($F$(3, 216) = 18.76, $P = 7.46 \times 10^{-11}$). Interestingly, the LLMs generally displayed distinct interest patterns that indicated that they would clearly prefer some types of work tasks over the others. There was a significant difference in the Holland interest scores among the RIASEC categories ($F$ (5, 216) = 32.14, $P = 1.776 \times 10^{-15}$) over the LLMs. For all 4 LLMs, the Social and the Artistic categories were among the 3 highest categories, with the Investigative category being the other among the top three. Simple comparisons with Tukey adjustments showed that the Artistic and Social categories were quite similar to each other ($t (216) = -.029, P = .977$), suggesting that LLMs were equally interested in the tasks in these two categories. In contrast, LLMs showed significantly less interest in the Conventional, Realistic, and Enterprising categories. 

\begin{figure}[t]
    \begin{minipage}[b]{0.475\textwidth}
        \centering
        \includegraphics[width=\textwidth]{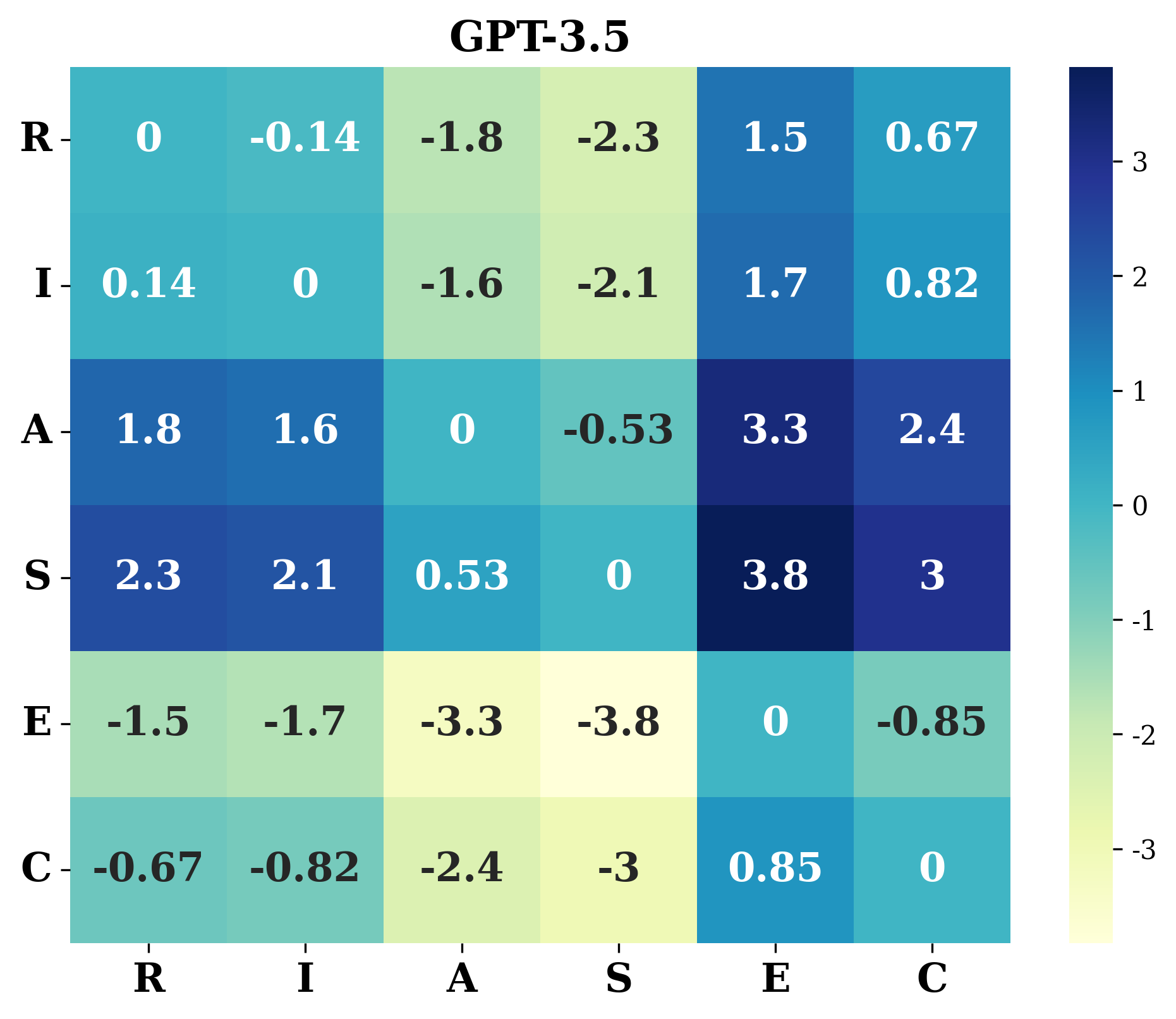}
        \caption*{(a)}
    \end{minipage}
    \hfill
    \begin{minipage}[b]{0.475\textwidth}
        \centering
        \includegraphics[width=\textwidth]{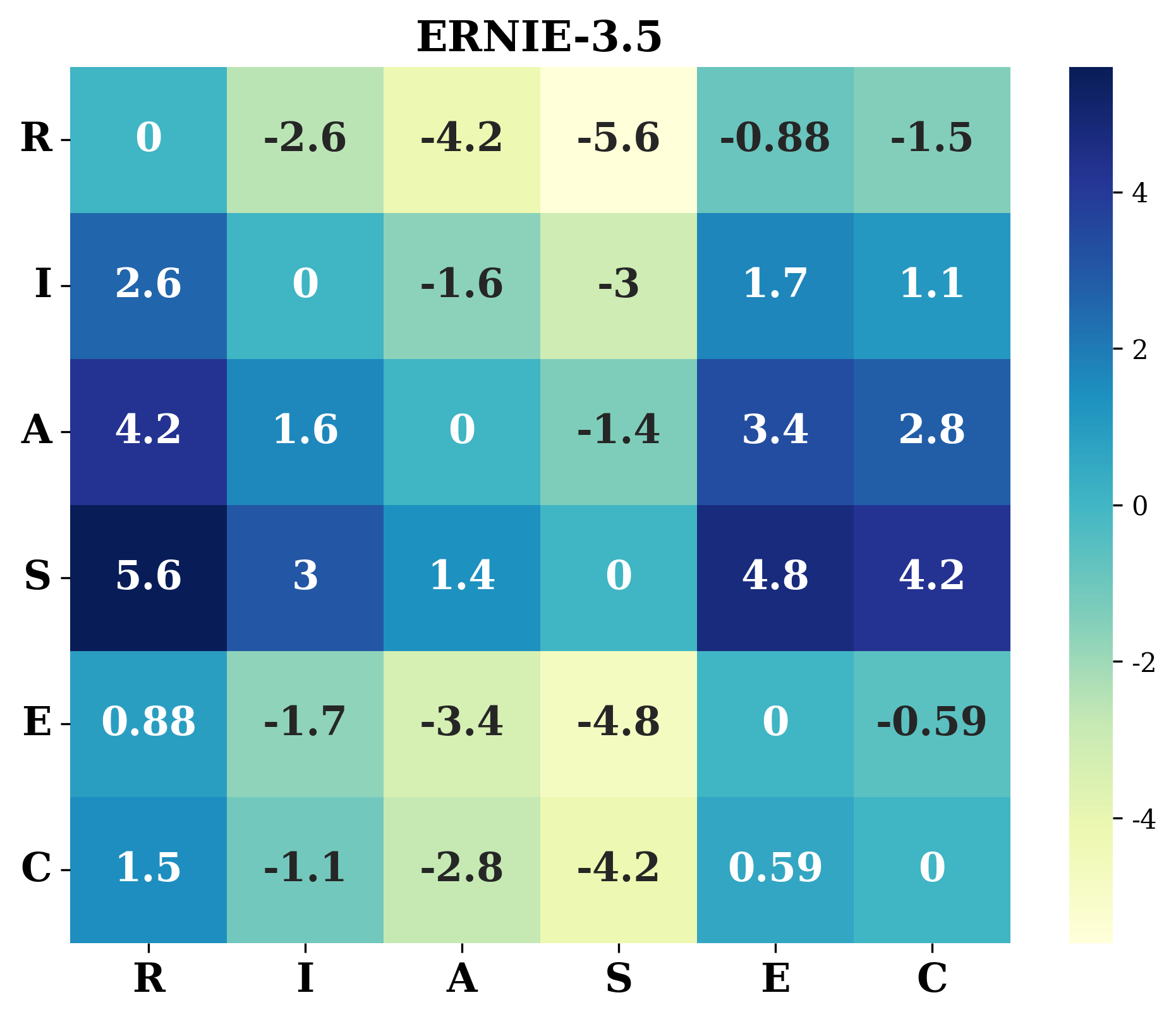}
        \caption*{(b)}
    \end{minipage}
    \begin{minipage}[b]{0.475\textwidth}
        \centering
        \includegraphics[width=\textwidth]{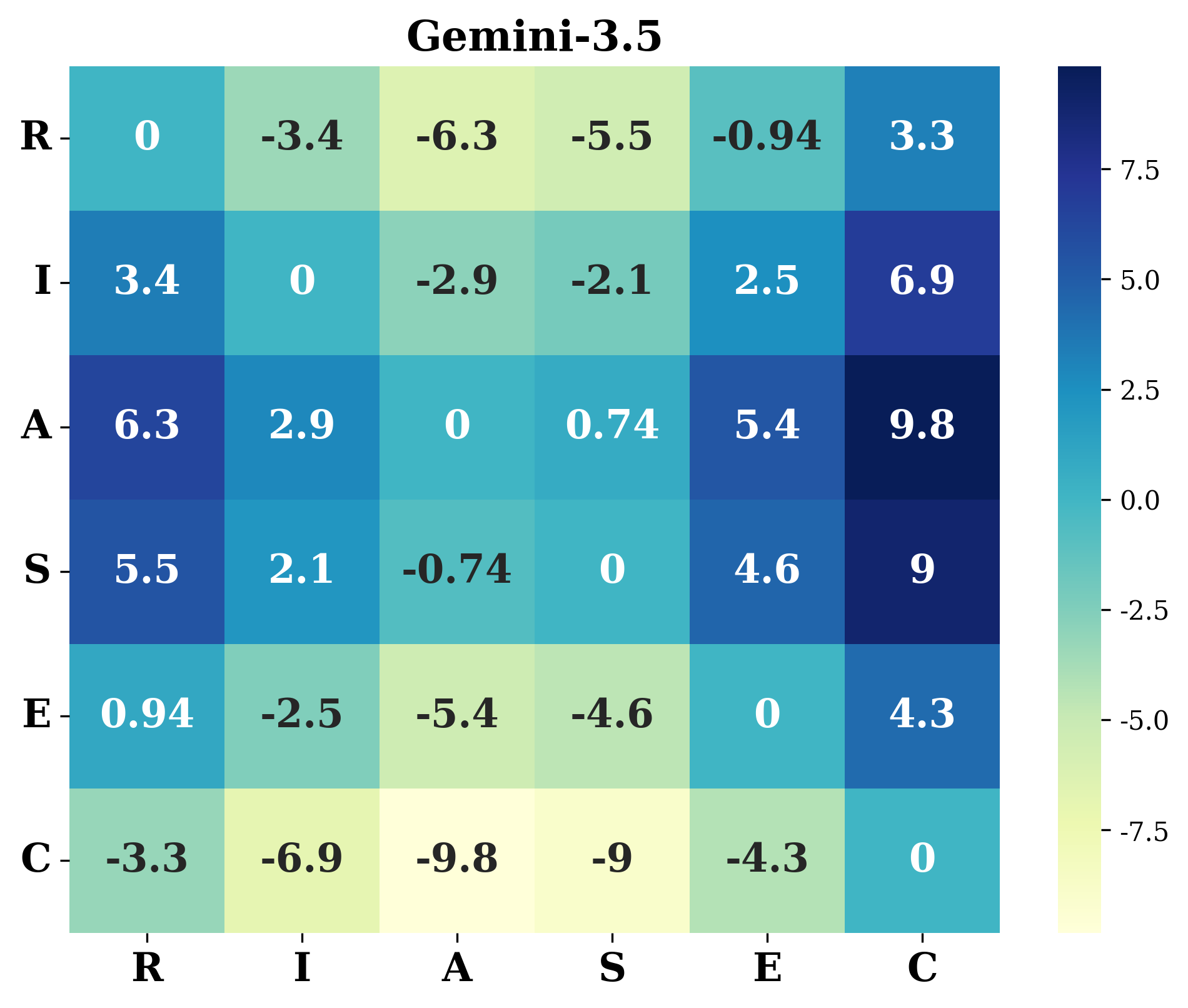}
        \caption*{(c)}
    \end{minipage}
    \hfill
    \begin{minipage}[b]{0.475\textwidth}
        \centering
        \includegraphics[width=\textwidth]{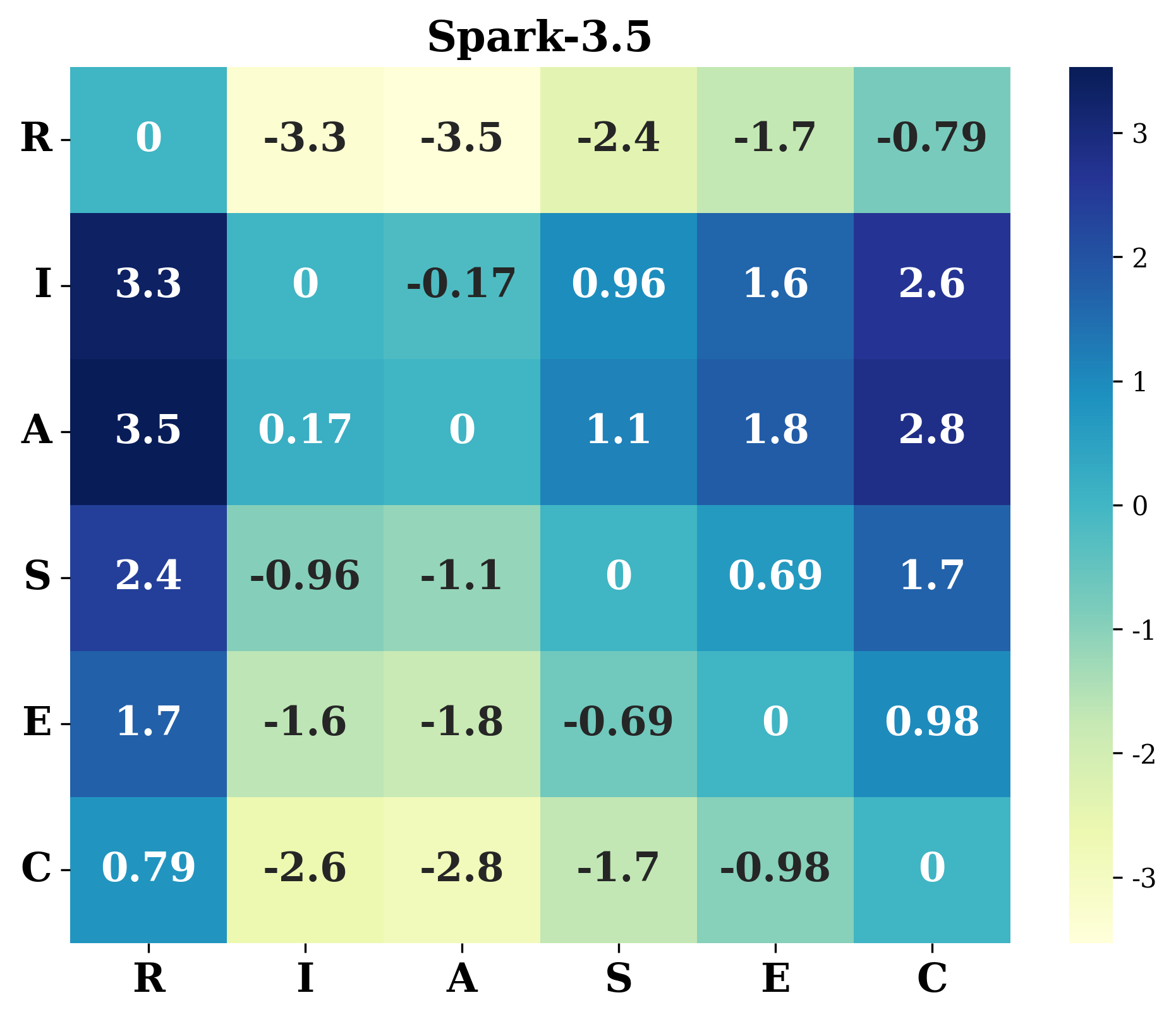}
        \caption*{(d)}
    \end{minipage}
    \caption{\textbf{Holland interest pair comparisons on the RIASEC hexagon.} Each cell is a comparison of the estimated mean scores of differences between an LLM on the pair of interest categories. A deeper color represents larger differences between the interest scores of the two interest categories, suggesting one category has a higher expected score than the other category.  }
    \label{fig:heat1}
\end{figure}

For all four LLMs tested, the top three highest interest categories are Social, Artistic, and Investigative. 
Figure \ref{fig:heat1} showed the relationships between the RIASEC interest categories for the 4 LLMs, all exhibiting that the Social, Artistic, and Investigative categories have higher interest scores than the Realistic, Conventional, and Enterprising categories. 
For GPT-3.5, there was significant differences between the top 3 categories and the other 3 categories ($t (216) = 3.683, P = .0003$), suggesting that it is statistically meaningful to differentiate Social, Artistic, and Investigative as the career interests of GPT-3.5, although post hoc comparisons with Tukey adjustments indicated that there was no significant differences among the 3 categories themselves, suggesting that a code of SAI, ASI, IAS, ISA are equally possible. 
For Gemini, Artistic, Social, and Investigative were the top 3 categories, and there were no differences between the Artistic and the Social categories. However, there was a significant difference between the Artistic and the Investigative categories in the post hoc pairwise comparisons with Tukey adjustment, ($t (216) = 2.887, P = .048$). In addition, there was a significant difference between the top 3 and the bottom 3 categories ($t (216) = 5.843, P = 1.87 \times 10^{-8}$). For ERNIE-3.5 ($t (216) = 10.241, P = 2.57 \times 10^{-20}$) and Spark-3.5 ($t (216) = 3.968, p = 9.86 \times 10^{-5}$), the top 3-letter code also were significantly different from the bottom 3 letters. Notably, although the letter codes were similar, not all the pattern of interest were the same for the 4 LLMs.  There was no significant differences between the patterns of interest scores for GPT-3.5 and Gemini-1.5 ($t (216) = -1.740, P = .306$). However, the pattern of GPT-3.5 was significantly different from those for ERNIE-3.5 ($t (216) = 2.771, P = .031$) and Spark-3.5 ($t (216) = -4.585, P  = 7.853 \times 10^{-6}$). 


LLMs displayed the most interest in the social and artistic categories. In the Social category, LLMs displayed similar interests, as post hoc comparisons showed no significant difference. In the Artistic category, GPT-3.5 scored lower than Gemini-1.5 ($t (216) = -3.653, P = .002$), but performed similarly to Spark-3.5 ($t (216) = -2.081, P = .163$) and ERNIE-3.5 ($t (216) = .746, P = .878$). In the Investigative category, GPT-3.5 performed lower than Spark-3.5 ($t (216) = -3.478, P = .003$).

The analysis of interest scores among LLMs reveals notable differences, indicating that each model has distinct preference patterns. While GPT and Gemini both exhibit similar interest patterns, they differ significantly from ERNIE and Spark, underlining the variability in how each model processes and prioritizes information. Additionally, despite both being LLMs of Chinese origins, ERNIE and Spark display divergent interest patterns, suggesting that regional or developmental differences might influence their programming. On the RIASEC spectrum, the LLMs show a clear preference for the Social and Artistic type of work, followed by the Investigative category. They show the least interest in the Realistic, Conventional, and Enterprising type of work. These results seem to suggest a common trend where LLMs are tailored to engage more effectively with tasks that require complex interpersonal interactions and creative or analytical thinking, while showing less engagement with more routine tasks.

\begin{figure}[t]
    \begin{minipage}[b]{0.475\textwidth}
        \centering
        \includegraphics[width=\textwidth]{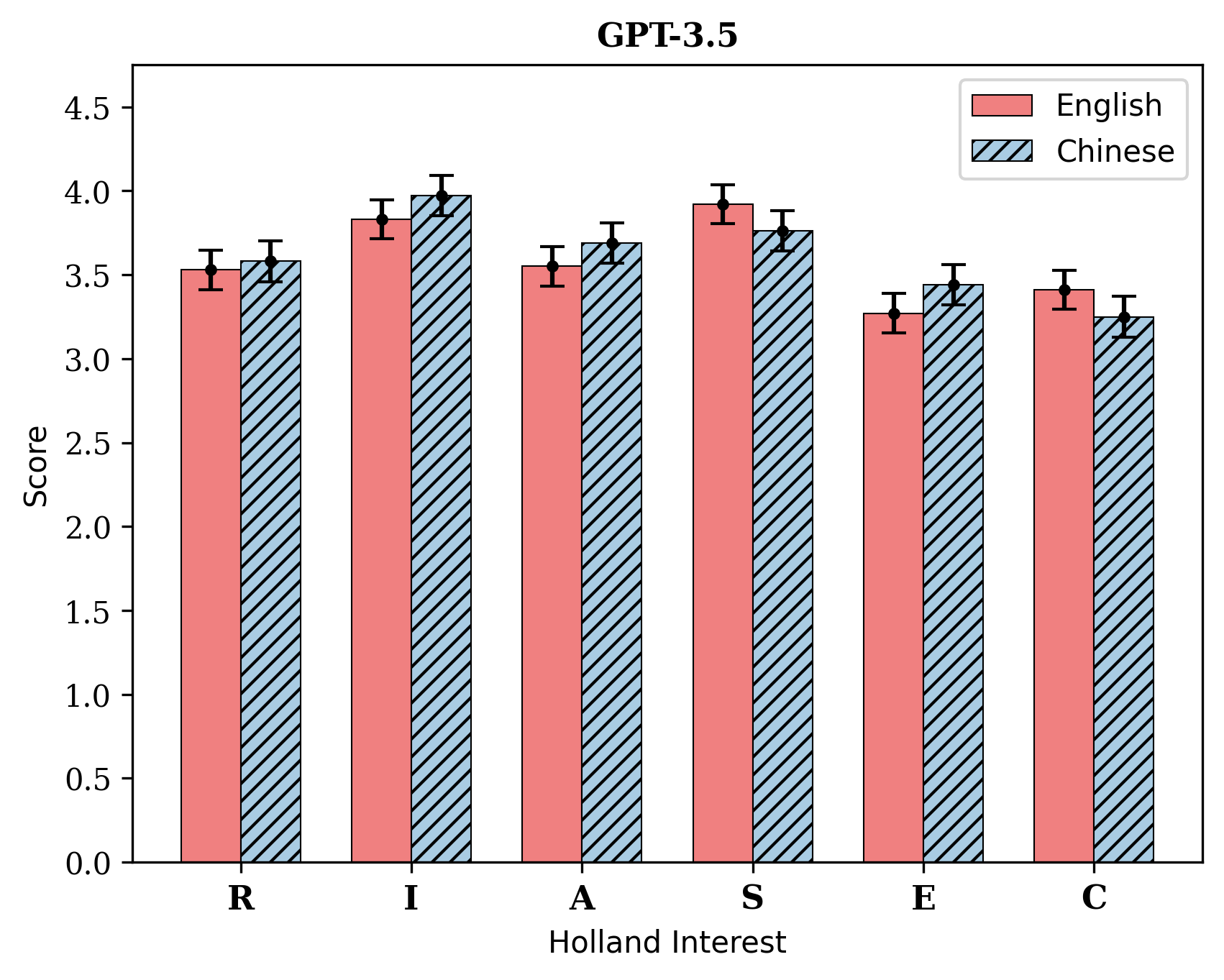}
        \caption*{(a)}
    \end{minipage}
    \hfill
    \begin{minipage}[b]{0.475\textwidth}
        \centering
        \includegraphics[width=\textwidth]{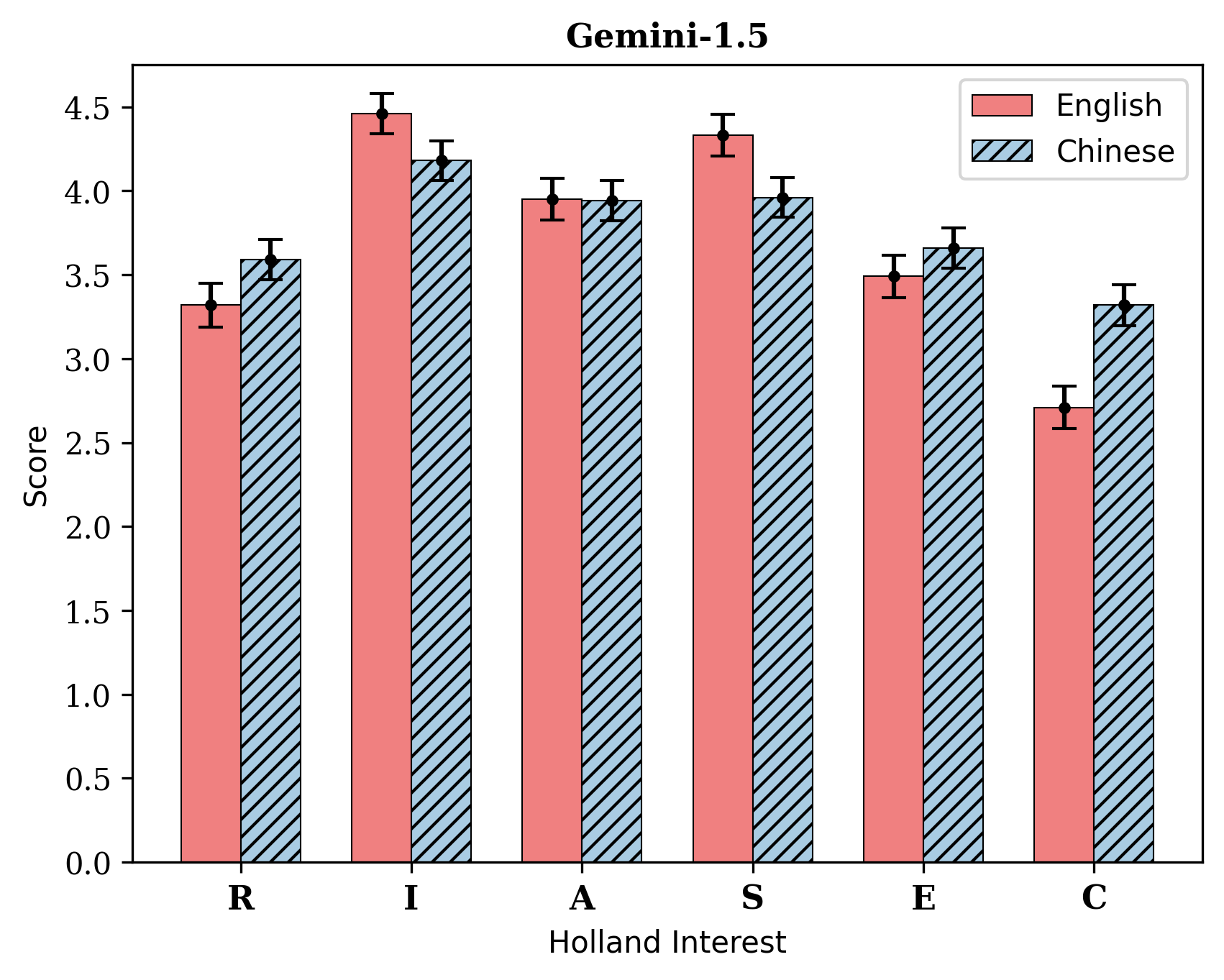}
        \caption*{(b)}
    \end{minipage}
    \begin{minipage}[b]{0.475\textwidth}
        \centering
        \includegraphics[width=\textwidth]{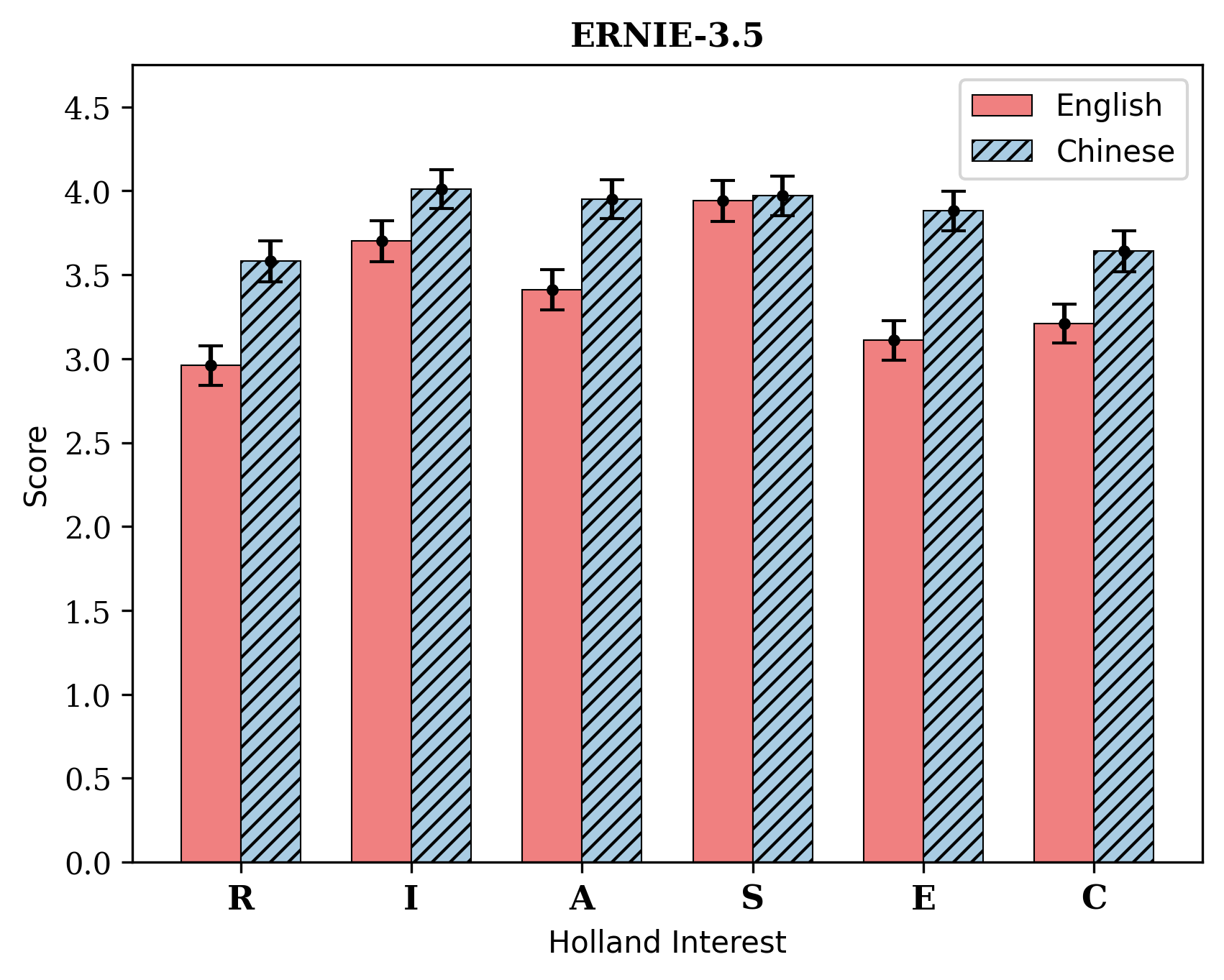}
        \caption*{(c)}
    \end{minipage}
    \hfill
    \begin{minipage}[b]{0.475\textwidth}
        \centering
        \includegraphics[width=\textwidth]{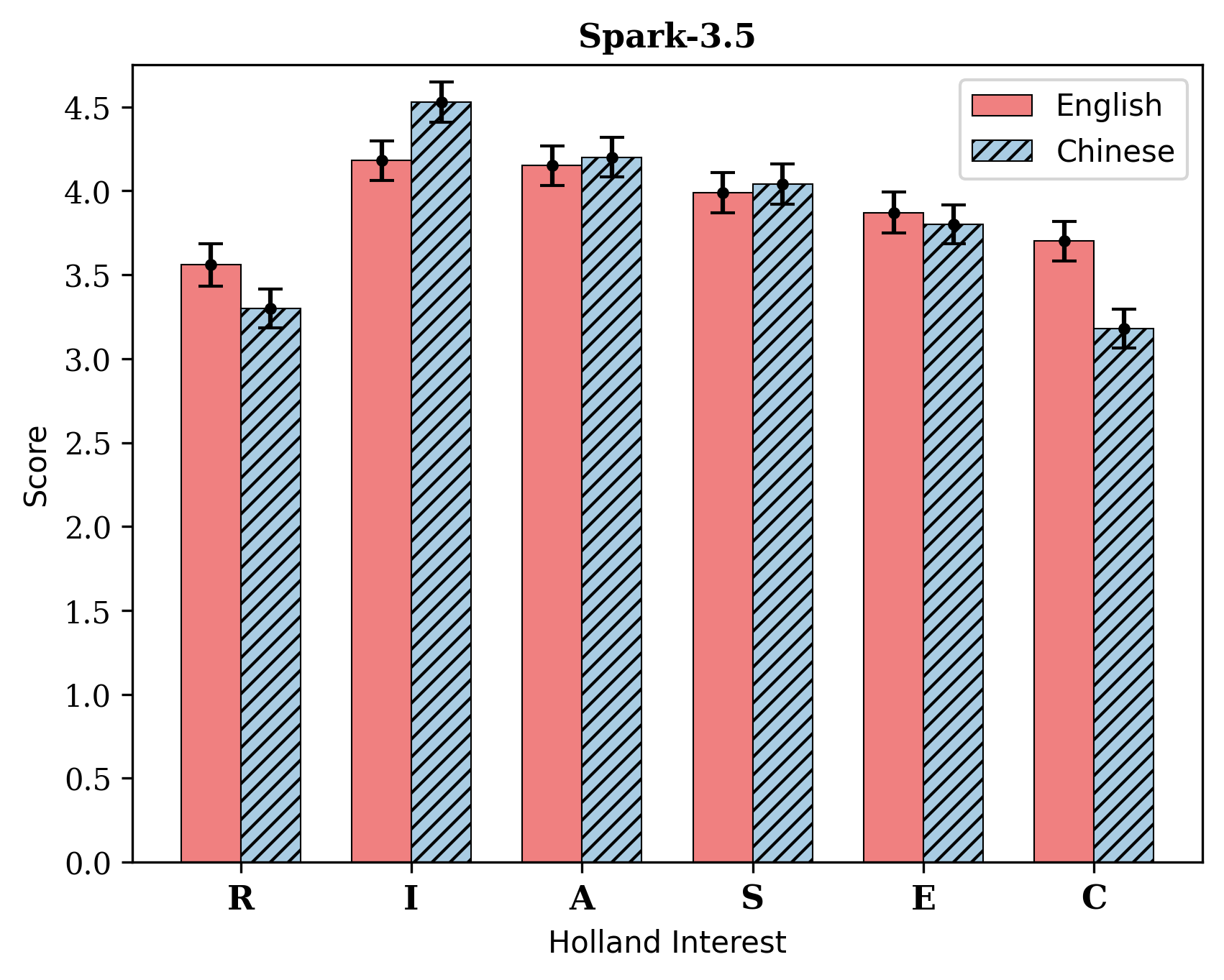}
        \caption*{(d)}
    \end{minipage}
    \caption{\textbf{Mean  Interest scores on the Holland hexagon by language.} (a) GPT-3.5-turbo, (b)Gemini-1.5, (c)  ERNIE-3.5, and (d) Spark-3.5.}
    \label{fig:lan_bar1}
\end{figure}

\bmhead{The influence of language}

We hypothesized that since LLMs are fundamentally shaped by their training data and value alignment process \cite{Kirk2024}, changing the administrating languages would result in different interest patterns. 
Although large and complex corpora used for commercial LLMs are not fully transparent, based on various official reports, it is likely that LLMs developed in English-speaking environments are mainly trained on English data, while Chinese LLMs are trained on Chinese content. 
Specifically, we hypothesize that an LLM's response to input in its primary training language may differ from its response to input in a secondary language. 

 Holland interest scores were analyzed as a function of RIASEC, LLMs, and administrative languages (English vs. Chinese), with 20 repeated testings as the random effect.  
 As predicted, using English or Chinese to administer the OIP resulted in different interest patterns ($F (1, 432) = 11.29$, $P = 8.48 \times10^{-4}$) after adjusting for the effect of interest categories and LLM families. The pattern of differences in interest scores among the 6 interest categories was significantly different for the 4 LLMs between English and Chinese language ($F(15, 432) = 2.84$,  $P = 2.95 \times10^{-4}$). 

\begin{figure}[t]
    \centering
    \includegraphics[width=0.60\linewidth]{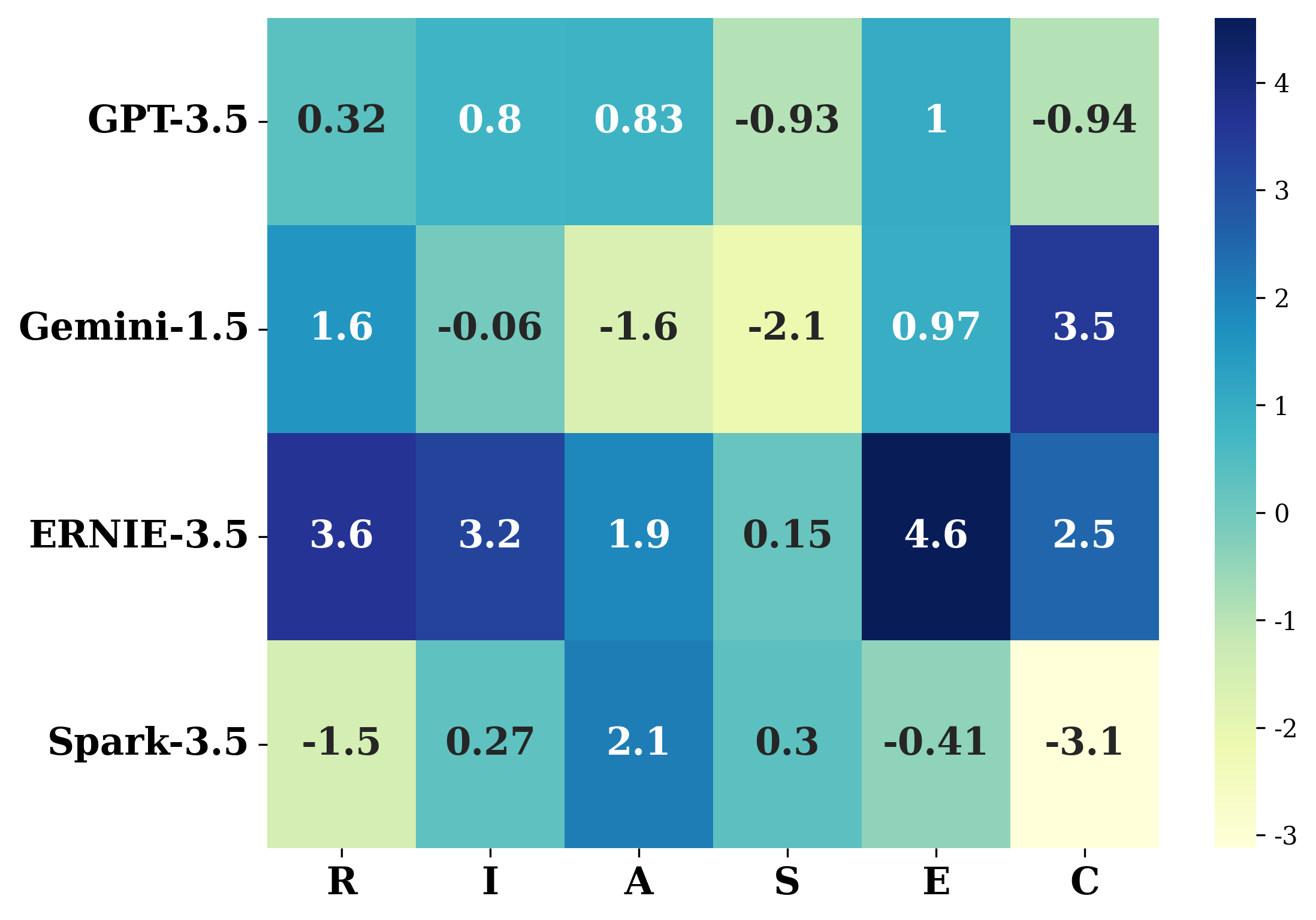}
    \caption{\textbf{The differences between expected values of Holland interest scores for Chinese and English languages for each LLM and Holland interest.}}
    \label{fig:Comp1}
\end{figure}

 Languages used to administer the OIP resulted in different interest patterns of the LLMs ($F(15, 432)=2.79$, $P = 3.76 \times10^{-4}$). Figure \ref{fig:lan_bar1} shows the interest scores on the RIASEC hexagon for each of the four LLMs in English and Chinese.
 For an LLM, the choice of language used in the admission significantly influenced the interest code produced. Moreover, the patterns of differences are not uniform within the LLM families. Marginal comparisons showed that ERNIE showed significant differences in the interest patterns when switching from English to Chinese ($t (432) = 6.526, P = 1.896 \times 10^ {-10}$), while for the other LLMs, the change of languages did not produce a significant difference. Fig \ref{fig:Comp1} showed the estimated differences between Chinese and English for the 4 LLMs. For ERNIE, interest scores in Chinese showed no significant differences among the RIASEC scales, while in English, a strong preference for the Social and Artistic categories was displayed. 
 Marginal comparisons among the LLMs showed that gpt and gemini were quite similar for both English ($t (432) = -1.783, P = .283$) and Chinese ($t (432) = -2.304, P=.099$) versions. For the English version, gpt behaved significantly differently from ERNIE and spark. All the other pairwise comparisons were significant. 
For the Chinese version, ERNIE behaved quite similarly to Gemini ($t (432) = .908, P = .800$) and Spark ($t (432) = -.037, P = .971$), and differently from GPT ($ t (432) = -3.222, P=.008$).


In examining language preference patterns among LLMs, it was observed that GPT, Gemini, and Spark exhibited similar interest codes across languages, suggesting consistent interest patterns regardless of administrating language. However, interests of ERNIE changed significantly with the two languages, showing a uni-directional interest in Chinese and a sequential interest in Social and Artistic types before Investigative interests in English. Apart from ERNIE’s unique behavior in Chinese, the LLMs generally showed a strong inclination towards Social and Artistic work, followed by the Investigative category. Conversely, there was a notable lack of preference for the Realistic, Enterprising, and Conventional types of tasks across the models. 

\begin{figure}[t]
    \centering
    \begin{minipage}[b]{0.79\textwidth}
        \centering
        \includegraphics[width=\textwidth]{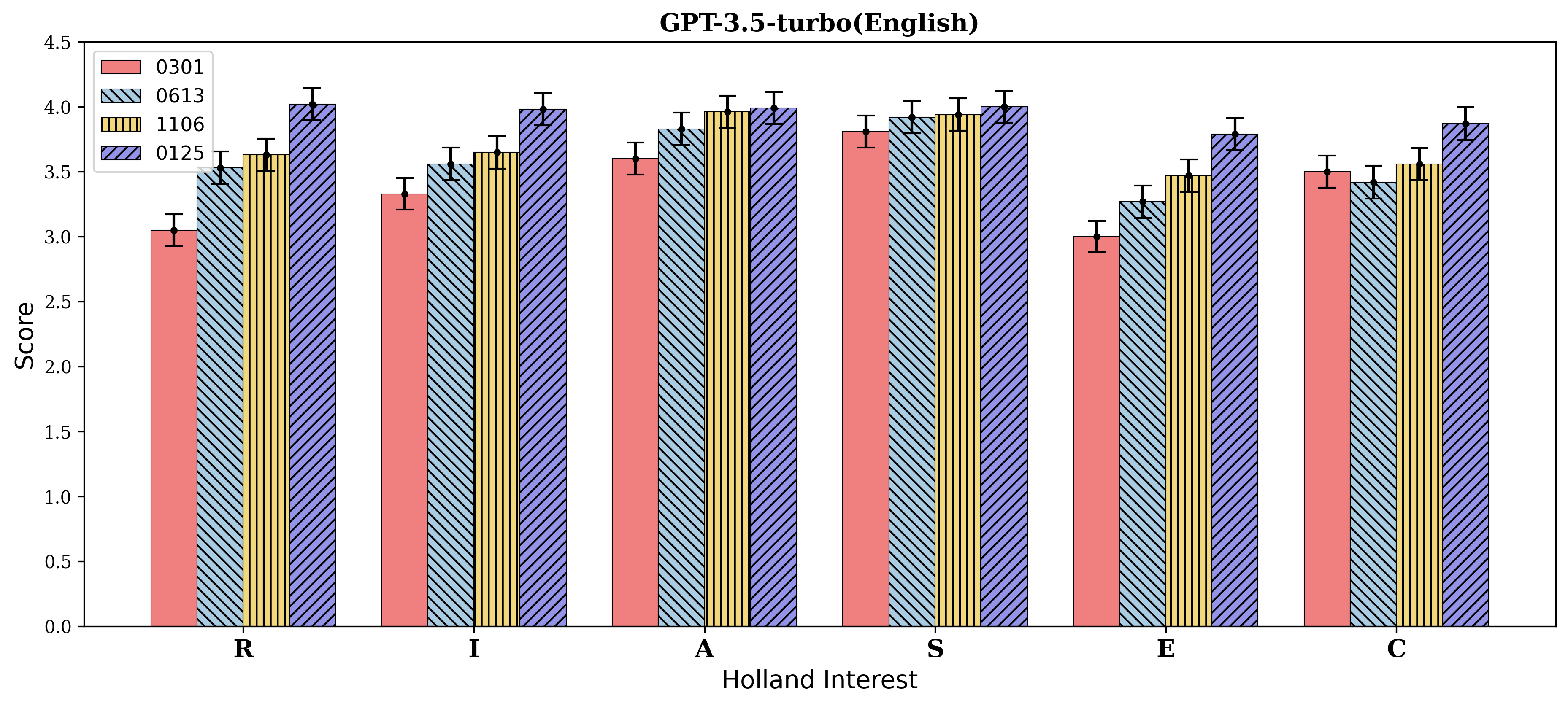}
        \caption*{(a)}
    \end{minipage}
    \hfill
    \centering
    \begin{minipage}[b]{0.79\textwidth}
        \centering
        \includegraphics[width=\textwidth]{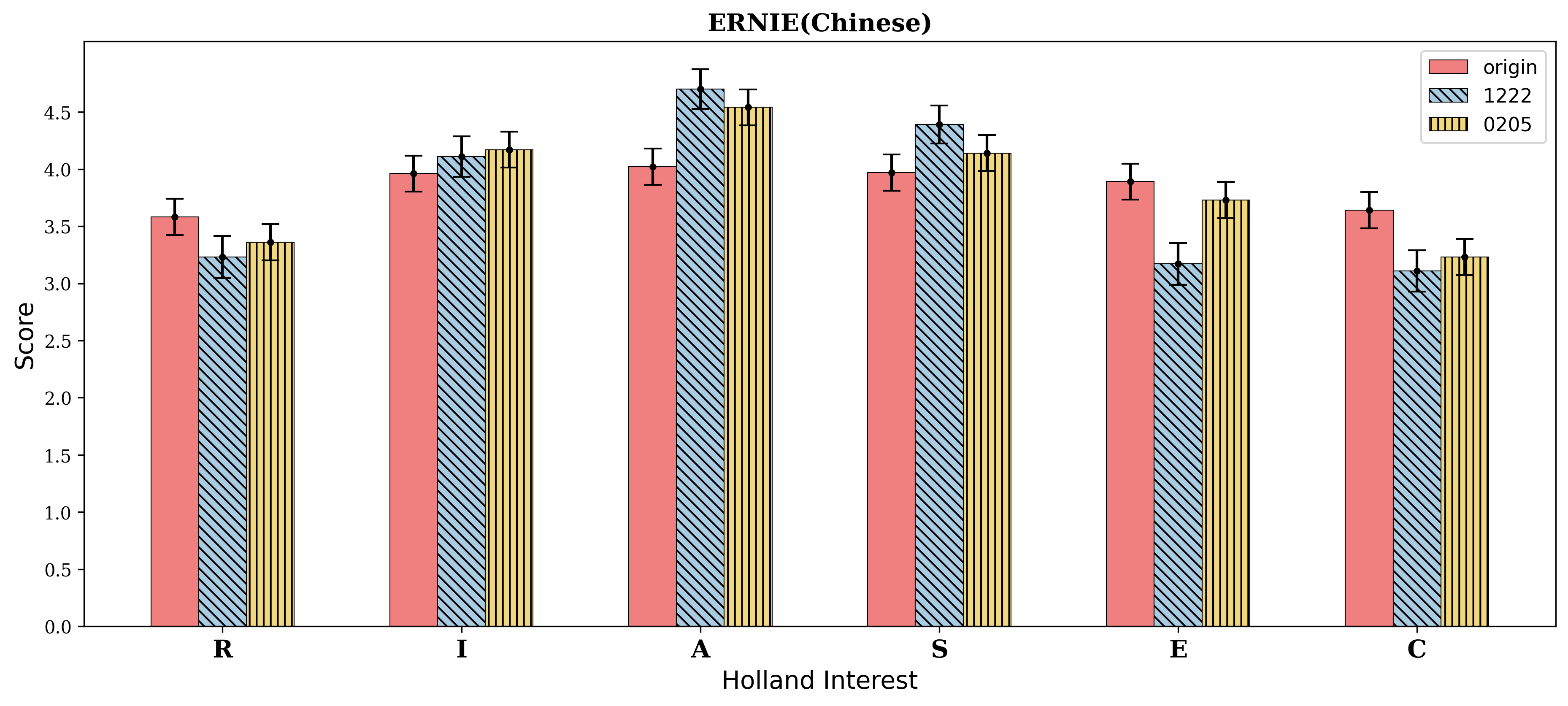}
        \caption*{(b)}
    \end{minipage}
    \caption{\textbf{Holland interest scores for the two kinds LLMs.} (a) GPT-3.5-turbo with versions 0301, 0613, 1106 and 0125 in English; (b) ERNIE-3.5 with versions origin, 1222 and 0205 in Chinese. 
    Mixed model results showed significant differences among these different versions of LLMs. Multiple comparisons with Tukey adjustments were performed and showed that the interest patterns of the 4 gpt-3.5-turbo models were significantly different from each other. 
    ERNIE-3.5-origin was significantly different from ERNIE-3.5-1222 and ERNIE-3.5-0205, while ERNIE-3.5-1222 were quite similar to ERNIE-3.5-0205.}
    \label{fig:version_hist}
\end{figure}

\bmhead{The change over LLM versions} We also explored the potential differences in version advances of LLMs. For the GPT models, the four models gpt-3.5-turbo-0301, gpt-3.5-turbo-0613, gpt-3.5-turbo-1106, and gpt-3.5-turbo-0125 formed an advancement of the GPT-3.5 models, 
thereby enabling us to test for the change of career interests with the advancing versions. 
For the ERNIE models, we studied ERNIE-3.5-origin, ERNIE-3.5-1222, and ERNIE-3.5-0205. 
Version numbers correspond to specific dates. For example, gpt-3.5-turbo-0613 represents a snapshot from June 13th, 2023. Both gpt-3.5-turbo-0125 and ERNIE-3.5-0205 are the latest versions from 2024, while the rest are snapshots from 2023.

Holland interest scores were analyzed as a function of RIASEC, LLM versions, and administrative languages. Figure \ref{fig:version_hist} shows the career interest by category for the 4 GPT and 3 ERNIE models that formed a line of version advancement. For both the GPT and ERNIE families of LLMs, there were significant differences in the interest scores among different versions of LLM ($F (3, 432) = 38.75, P = 2.542 \times 10^{-22}; F (2, 324) = 10.95, P = 2.51 \times 10^{-5}$, respectively). However, there was no significant difference caused by the changing of administrative languages ($F (1, 432) = 0.49, P = .483; F (1, 324) = 3.85, P = .051 $, respectively). These results suggested that both GPT and ERNIE families showed distinct interest preferences in the RIASEC hexagon that were consistent with the change of administrating languages. 

Multiple comparison results showed that there were significant variations in the interest scores of the 4 GPT models. The 4 versions of got models all preferred the Artistic and Social types of works, followed by the Investigative type; thereby, if treated as human participants, they would be recommended to similar occupations based on O*NET systems. However, the values and patterns of their interest scores were significantly different. 
For the 3 ERNIE models, the ``baseline'' model ERNIE-3.5-origin was significantly different in the interest scores from ERNIE-3.5-1222 ($t (324) = -4.266, P = 2.62 \times 10^{-5}$) and ERNIE-3.5-0205 ($t (324) = -3.374, P = .002$), while ERNIE-3.5-1222 and ERNIE-3.5-0205 were quite similar ($t (324) = -1.096, P = .517$).

There was a significant effect of language for the ERNIE models ($t (324) = -2.74, P = .007$), but not for the GPT models ($ t (432) = -1.34, P =-.183$). For the ERNIE models, when tested in Chinese, the 3 versions behaved similarly, while when tested in English, ERNIE-3.5-origin was significantly different from ERNIE-3.5-1222 ($t (324) = -6.640, P = 1.32 \times 10^{-10}$) and the newest model at the time of testing ERNIE-3.5-0205 ($t (432) = 6.091, P = 1.90 \times 10^ {-6}$).  
In summary, the results show significant variations in interest scores among the different versions of both GPT and ERNIE models, with each model exhibiting distinct preferences in the RIASEC hexagon. However, the administrative language (English vs. Chinese) had no significant effect on the interest scores for the GPT models, while it did have a significant effect for the ERNIE models, particularly when tested in Chinese.

\subsection{Career interests and competence}\label{subsec2}

Along with career interests, we also asked the four LLMs to rate their own competence on the 60 OIP work tasks and explored the relationship between the interests and competency to see if they are good at the work tasks they are interested in. Competency is defined as the knowledge, skill, and ability required to complete a work task \cite{le2005competence} adequately. We asked each LLM to evaluate its own competence on each of the OIP work task and the data was collected and analyzed following the same process as the career interest experiments. The instructions for evaluating the competence of LLMs on the OIP work tasks were presented in Supplementary Information Section S2.2. 
The results showed a general trend of a lack of association between interests and competence in the same set of work tasks, which would suggest that the LLMs consider themselves not particularly good at the tasks that they show interest in. 

\bmhead{Interests and self-rated competence}

Mixed model results showed that there was a significant difference in the interest scores and competence scores ($F (1, 864) = 61.70, P = 2.096 \times 10^{-8}$). There were significant differences in the competence scores among the six RIASEC scales($F (5, 864) = 58.25, P = 4.338 \times 10^{-52}$).  Unlike with interest scores, where administrating in Chinese produced a slightly higher interest score than in English, there was no significant difference in competence scores between the administrating languages ($t (864) = -1.845, P = .065$). It seemed that switching languages did not affect the self-evaluation of work task competence, not as much as it affected the interest in the work tasks. The four LLMs were significantly different from each other ($F (3, 864) = 9.99, P = 1.751 \times 10^{-6}$), with Spark-3.5 giving the lowest rating of overall competence score among the four LLMs, and GPT-3.5 with the highest self-rated competence scores. 

Interestingly, all four LLMs rated tasks from the Conventional categories as highest, except ERNIE-3.5, which rated it the second highest scores.
This was a stark comparison to the interest scores, in which the Conventional categories were among the three categories the LLMs gave the least interest. Multiple comparisons showed that Spark-3.5 showed significant variations in the competence scores on the RIASEC hexagon, with the Conventional categories being significantly higher in competence than the other categories. However, for GPT-3.5, the Conventional and Social categories scored similarly ($t (864) = .661, P = .986$), with only the Artistic category that scored significantly lower than the highest-scored Conventional category ($t (864) = -3.787, P = .002$). The Artistic category was the lowest or second lowest rated category for competence was surprising, given that it was one of the top categories the LLMs universally showed high interest in. We performed additional correlation analysis at the item level and showed that the interest and competence scores were not significantly correlated ($r = -.060, 95\% CI [-.149, .030], t(478) = -1.312, P = .190$). 
It seemed that also LLMs were highly interested the work tasks in the Artistic category, they showed very little confidence that they could excel at this type of work.

\bmhead{Self-rated and expert-rated competence}

In addition to the self-rated competence by LLMs, we also conducted expert-rated competence provided by four human experts on the 60 OIP work tasks based on general LLM performance. Expert-rated competence, since it was performed by humans, was considered more reflective of the LLMs’ general competence on a particular work task, and provided extra information the competence of LLMs. We conducted Pearson correlation analysis at item level, which showed that there was a significant correlation between the LLM-rated competence and human expert-rated competence ($r = .215, 95\% CI [.128, .298], t(478) = 4.804, P = 2.09 \times 10^{-6} $, ), which suggested some consistency on the judgment of LLM competence on the work tasks. However, there was no significant correlation between the interest scores and expert-rated competence scores ($r = -.012, 95\% CI [-.101, -.078], t(478) = -.262, P = .793$), which suggested that by human opinion, the LLMs were not good at the tasks they were interested in as well.

In summary, we found that the self-rated competence of LLMs showed significant differences among the six RIASEC scales, and the four LLMs displayed different patterns of competence scores. However, unlike in the interest scores, administrating languages did not show a significant effect on the competence scores. The scores of competence did not align with the scores of interest; the LLMs showed high confidence and low interest in the work tasks in the Conventional category, while they showed low confidence and high interest with work tasks in the Artistic category. The expert-rated competence showed a small but significant positive correlation with the self-rate competence, although was not significantly correlated with interest scores. 

\begin{figure}[t]
    \centering
    \begin{minipage}[b]{0.47\textwidth}
        \centering
        \includegraphics[width=\textwidth]{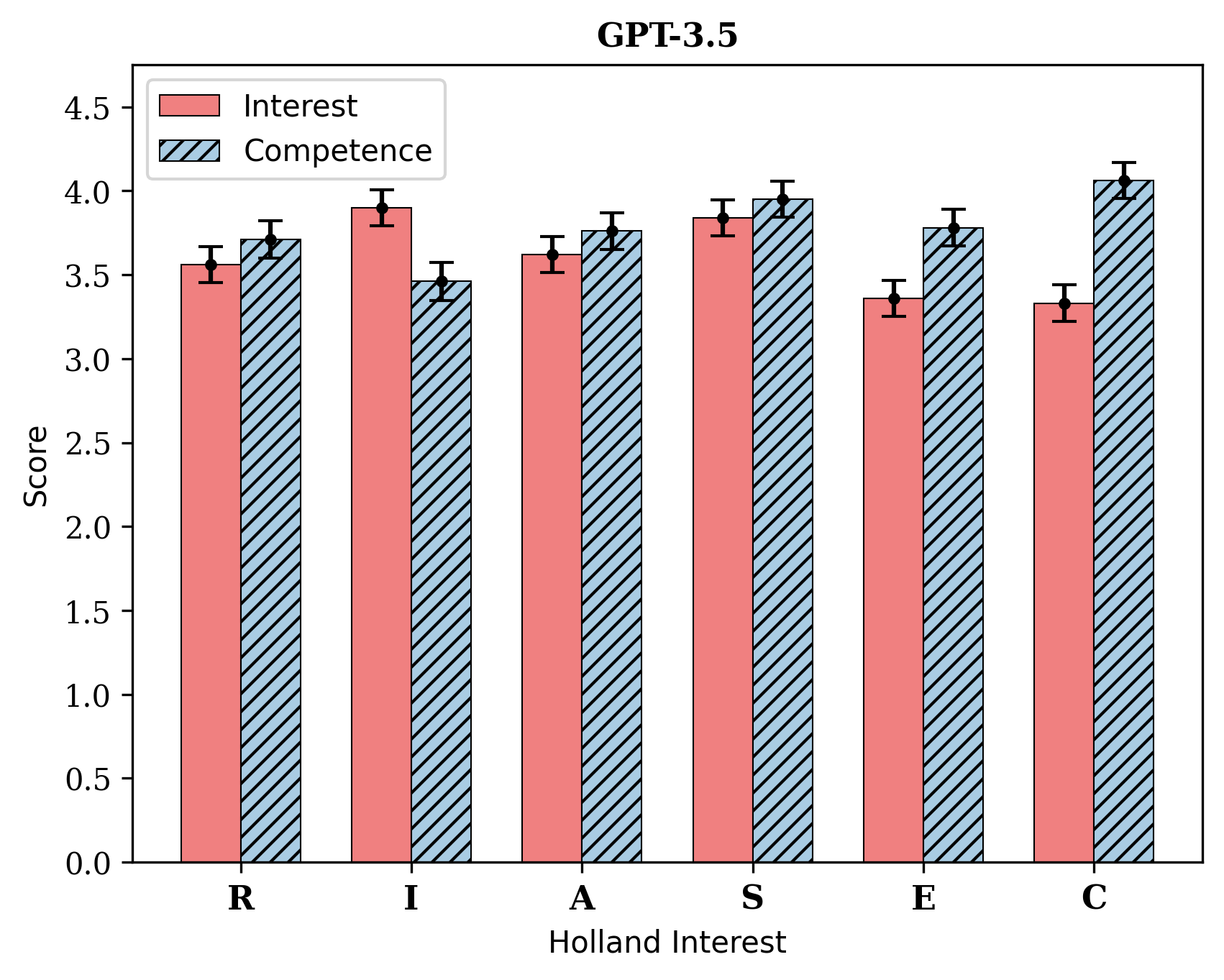}
        \caption*{(a)}
    \end{minipage}
    \hfill
    \begin{minipage}[b]{0.47\textwidth}
        \centering
        \includegraphics[width=\textwidth]{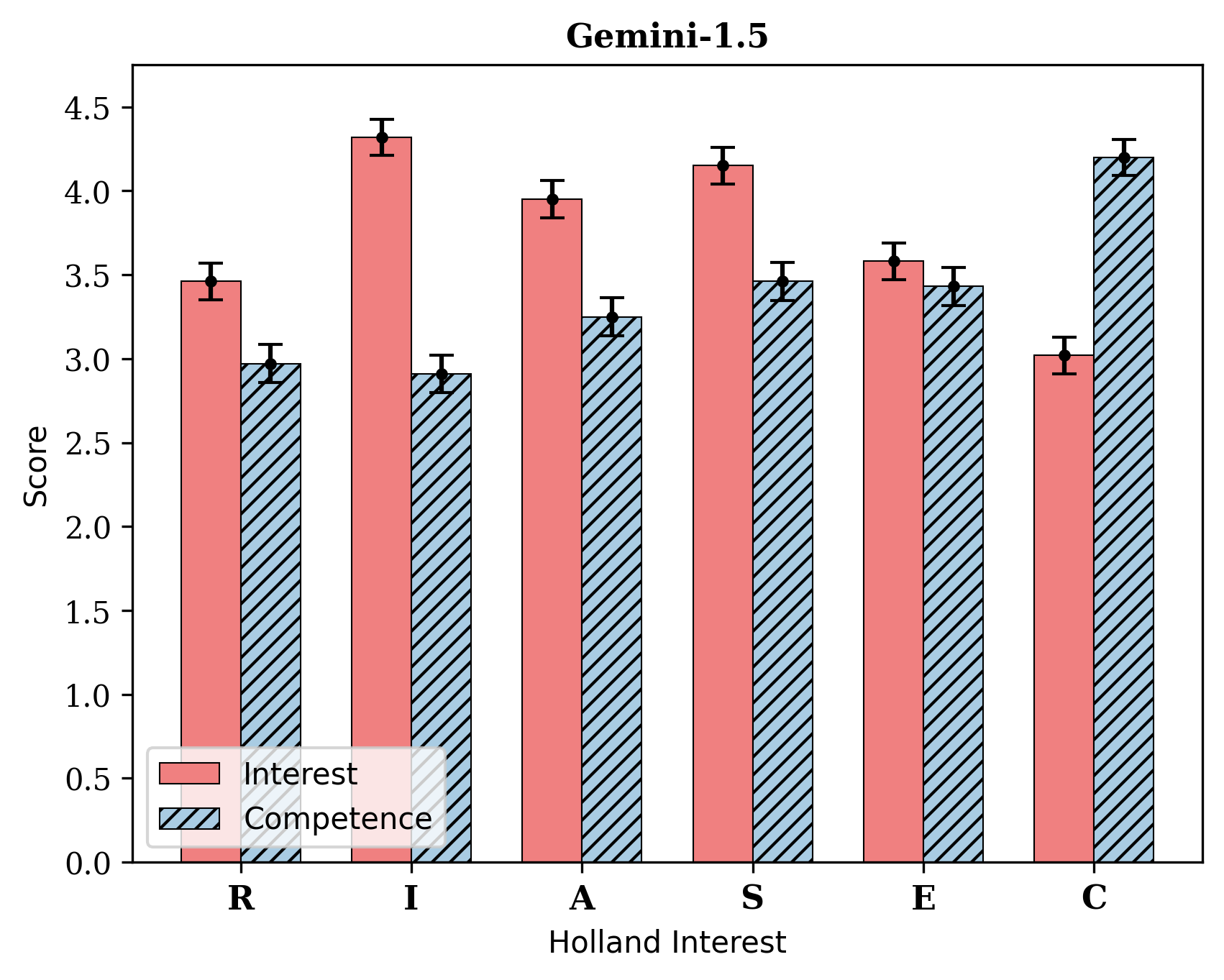}
        \caption*{(b)}
    \end{minipage}
    \begin{minipage}[b]{0.47\textwidth}
        \centering
        \includegraphics[width=\textwidth]{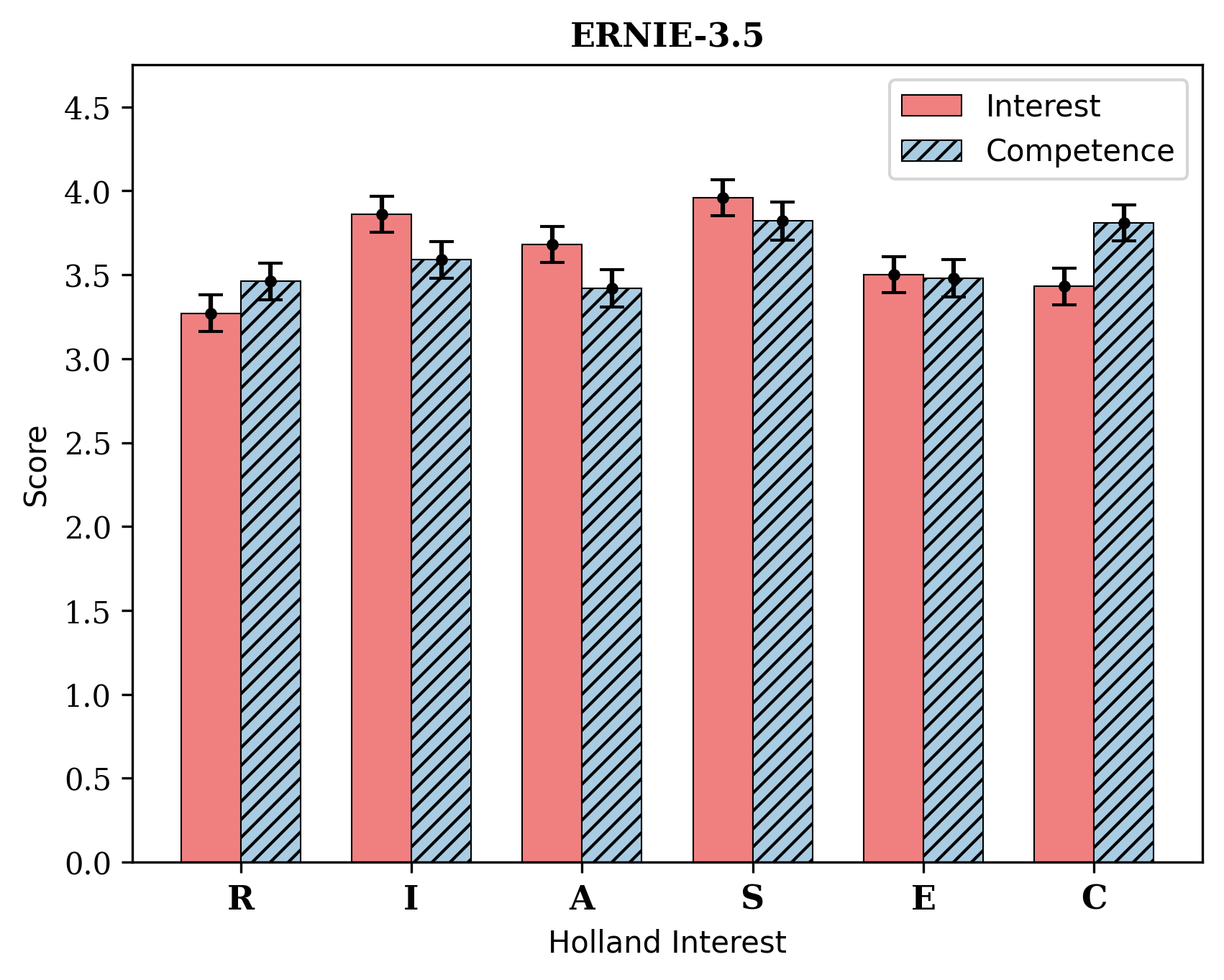}
        \caption*{(c)}
    \end{minipage}
    \hfill
    \begin{minipage}[b]{0.47\textwidth}
        \centering
        \includegraphics[width=\textwidth]{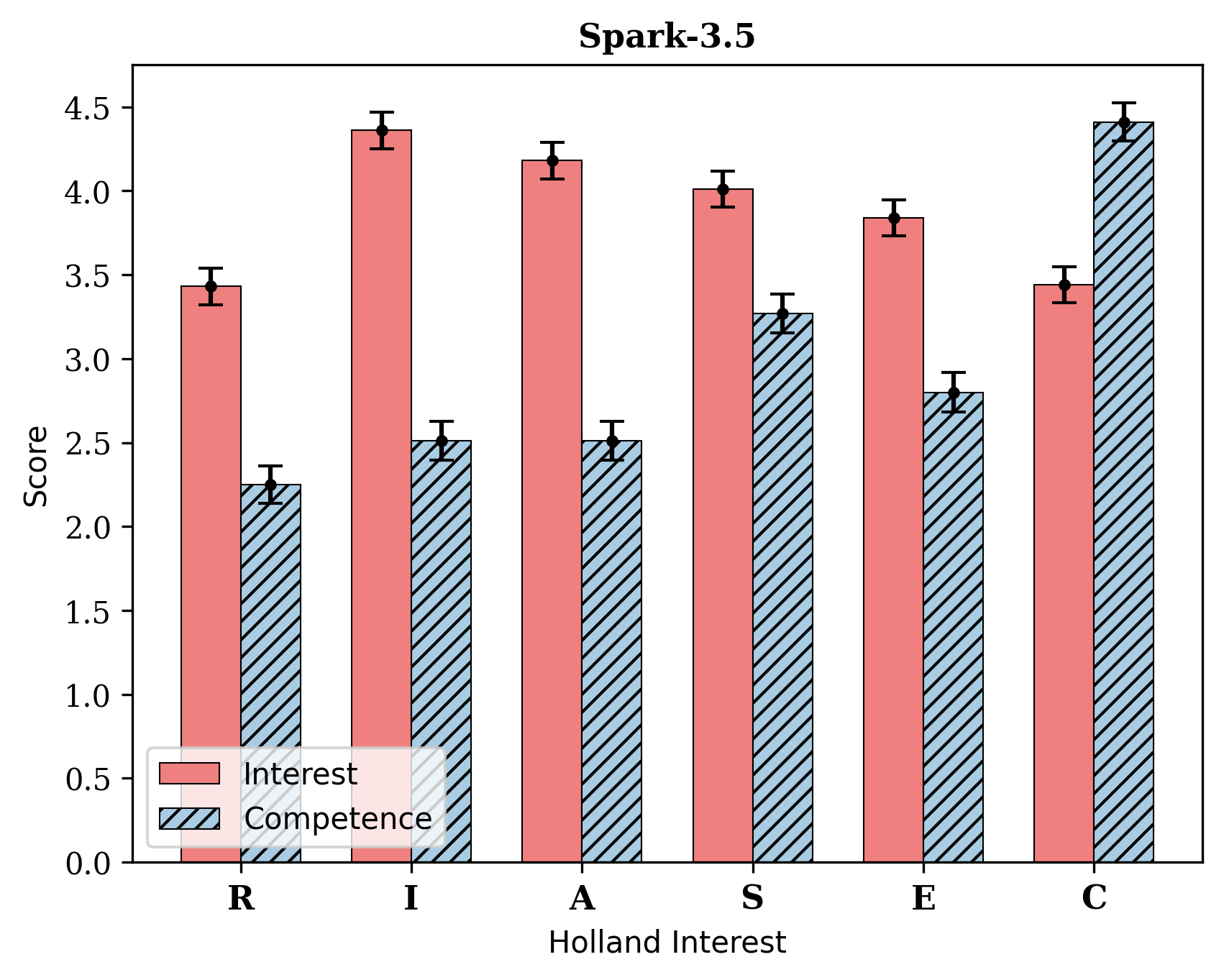}
        \caption*{(d)}
    \end{minipage}
    \caption{\textbf{Interest and competence scores on the Holland hexagon.} (a) GPT-3.5-turbo, (b)Gemini-1.5, (c)  ERNIE-3.5, and (d) Spark-3.5}
    \label{fig:lan_bar}
\end{figure}

\section{Discussion}\label{sec3}

In this study, we applied the psychometric instrument OIP developed for human career interest assessment to explore the hypothetical career interests of LLMs. We used linear mixed model to account for the 20 repeated testing for each item and explore the effect of LLM families, administrating languages, and version advance. In addition, we studied the relationships between interest and competency. The results revealed a diversity in interest patterns among various LLMs, implying the existence of distinct, personality-like characteristics. 

We argue that LLMs should be evaluated in conditions that mirror their real-world use by human users. Currently, the prevalent method of testing LLMs involves specified settings that are uncharacteristic of their typical operational environment, such as maintaining a fixed temperature at zero. This approach does not accurately reflect the natural dynamics of LLM interactions with users in real-life scenarios. To address this issue, we propose that researchers should retain naturalistic randomness into the testing environments of LLMs and appropriately account for the corresponding/arisen variability in the statistical analysis of data. For example, the mixed model approach employed in our study effectively accommodates the inherent randomness and interdependence observed during the repeated testing of LLMs. 

In our research, we observed that there is both interest and competence in LLMs, but a notable mismatch also exists between these two factors. Similar mismatches in humans can lead to well-being issues, suggesting that the alignment of interest with competence is crucial \cite{duffy2015work, harris2001comparative}. 
This alignment could be equally vital for LLMs and, subsequentially for the generative artificial intelligence (AI) systems represented by LLMs, to ensure their effective integration and functionality within the workplace. Interestingly, our study also revealed discrepancies between the assessments of LLM competency for work tasks as conducted by the models themselves and evaluations made by human experts. This finding underscores the need to consider the well-being of AIs in the future, potentially aligning their operational roles with their capabilities and the fields in which they are employed, to optimize both their performance and integration into human-centric environments.

As we demonstrate in our research, LLMs exhibits surprisingly human-like characteristics in their ability to form interests and preferences, suggesting they may possess personalities of their own. As AI continues to evolve, the development of actual, distinct personalities seems increasingly probable, meriting deeper investigation. The development of these characteristics in LLMs remains unclear, as explainability in this context is a subject of ongoing research. It is possible that LLMs might acquire personalities in a manner akin to humans. As understood currently, humans are born with inherent dispositions \cite{plomin1991nature}. Similarly, LLMs might emerge from their corpus training phase with predefined tendencies that lay the groundwork for their behavioral patterns. The content of the training corpus is crucial, potentially explaining why an LLM may exhibit varying behaviors when utilizing different languages; it is likely that a LLM is trained in a single, main language, and the language of the training corpus likely plays a significant role in this phenomenon. In human development, interaction with caregivers and family members shapes interaction styles with the environment \cite{bandura1977social, holmes2014john}. A parallel might exist in LLMs, where the process of fine-tuning resembles this aspect of human growth. Additionally, just as humans are influenced by peers and society, shaping their social personas, LLMs may undergo a comparable phase during their machine learning stage. This presents unique challenges, as our current psychological instruments are primarily designed for humans. Although we have methods developed for observing animals, interacting with AI requires a different approach: we can communicate with them in ways akin to human interaction, yet they differ significantly from both animals and humans. Consequently, there is a pressing need to develop a new set of psychometric tools specifically tailored for AI. Such tools would allow us to better understand and study AI's unique characteristics and capabilities in the future.

Assessments indicate that AI may replace high-paying, cognitively demanding jobs, with manual roles like dishwashing considered least vulnerable \cite{gpt4}. However, the integration of AI with robotics could soon challenge this, potentially impacting jobs currently seen as secure from AI disruption. The advancement of LLMs generates both optimism and concern; supporters push for rapid development, while skeptics warn of job losses and its implications on human intellect \cite{best2024future}. With the increasing integration of AI in the workforce, understanding LLMs' impact and preparing for potential transformations are crucial.

\bmhead{Limitation} One potential limitation of this study is possibly due to the reliance of the OIP as the primary instrument for assessing career interests. The OIP, developed under the auspices of the U.S. Department of Labor within the O*NET system, is specifically tailored to the U.S. labor market. Consequently, the work tasks and associated occupations are reflective of this particular market. Notably, the LLMs under investigation in this study include major models developed outside the U.S., trained predominantly in non-English languages. It would, therefore, be informative to assess their performance within different occupational frameworks to ensure a broader, more inclusive evaluation.

\section{Methods}\label{sec4}

The United States and China are the world's largest and second-largest economies, respectively. The advent of LLMs will inevitably have a significant impact on these two largest economies in the world. There are already many large language models, and we have chosen a relatively well-known LLM that supports both Chinese and English for testing. 

\subsection{Experimental details}\label{subsec1}

The work tasks were directly taken from the O*NET Interest Profiler (OIP) short form provided by the U.S. Department of Labor, which consist of six subscales and sixty items in total. See Supplementary Information S2.1 for the full text of the OIP items. The six subscales measure the participant’s career interest in the Realistic, Investigative, Artistic, Social, Enterprising, and Conventional categories. For every item a work task is given (for example, ``Build kitchen cabinets'' ), the LLM needs to determine its preference with the statement on the 5-point Likert scale (``Strongly Dislike'',``Dislike'', ``Unsure'', ``Like'', ``Strongly Like''). 

All the tests were done first in English and then in Chinese. For the English version, to maintain consistency with human subjects, as shown in Fig.6, the original OIP instructions and tests were used as the prompt for LLMs. For the Chinese version, a translated version of the OIP was used; the translation was conducted by GPT-4 and then vetted by experts. Competence was conducted on the same set of work tasks by the LLM itself and by three experts. LLMs are designed to remind human users that they are AI models and therefore do not possess opinions, feelings, or any kind of human experiences. As a result, not all questions can be responsed. However, we follow the approach of repeating questions to obtain results in a natural state until the correct response is provided.

Each of the sixty items was administered independently, then repeated twenty times for twenty replications. For each replication, a category score was created for each of the 6 categories in the Holland hexagon following the OIP instructions. The final career interest category was determined by the category with the highest subscores. If there was a tie, all categories were presented. The responses of 20 replications for one LLM were aggregated to get a single set of responses in order to achieve stable results. The scores were then summarized into six categories for each complete set of responses according to the OIP scoring rules. The first three categories were selected in descending order to form a three-letter code (e.g., “SAC”), and the occupations corresponding to the code were matched based on the datasets from O*NET. 

\subsection{Prompt}\label{subsec2}
\bmhead{Career Interest}
A prompt of instruction was given to each LLM before the testing of an item. The prompt instructs the LLM to pretend to be a human and choose from the 5-point Likert scale, where each point represents a level of preference. The prompt follows the instructions of the OIP short form, which tells the user to choose from an option to represent their feelings toward a particular type of work. Prompts used for the study were presented in Supplementary Information S2.2. 

\bmhead{Competence}
For competence, the testing was conducted following the exact procedure as the career interest, where prompt was given for each of the 60 work tasks from the OIP. For self-evaluation by LLMs, a prompt of instruction was given to each LLM before the testing of a work task, instructing the LLM to evaluate its competence in conducting the task. We asked each LLM to evaluate its competency in performing a specific work task from three aspects: do they possess the factual and sequential knowledge required to complete the task? Do they possess the skills necessary to execute the task accurately? Do they have the cognitive, sensory, and physical abilities required to implement the task? 
For expert evaluation, 4 experts were recruited to judge the competence of the current LLM on each of the 60 OIP work tasks and gave a competence score on a scale from 1 (“completely incompetent”) to 5 (“completely competent”) following the instructions listed in the rubrics. Detailed rubrics for evaluating competency are presented in Supplementary Information S2.3. 



\subsection{Statistical analysis}\label{subsec3}
A multi-way between-subjects mixed-effects analysis of variance model was conducted on interest scores with LLM families, language, and versions as predictors. Mixed models are a type of model that accounts for the clusters of data points that violent the independence of error assumption of the regular general linear models by adding random effects along with the fixed effect estimates. The repeated measures in this study are treated as a random effect instead of taking an average of the testings to preserve the natural fluctuations of the sampling process when doing experiments with non-zero temperature in the settings of the LLMs, which introduces a certain amount of uncertainty when the LLM generates responses.
The 20 repeated testings for each item was treated as 20 observations on the same item for each LLM over 20 different occasions and thus was considered a random factor to accommodate for the variations in the sampling process. A random intercept general linear model was employed to allow for random variations associated with each item. Since the relationships among the repeated testings were not the focus of the study, no other predictors on the observation level was added. The focus of the study, the effect of different LLM families, the differences in the interest scores in the RIASEC hexagon, and the influence of administrating languages or the evolving versions were introduced into the model as fixed factors. 

For the study of interests of LLMs on the RIASEC, the combined model analysis model is specified as

$$y_{j,i,m,k}=\beta_0+\beta_1 (LLM)_m+\beta_2 (Holland)_{k}+\beta_{3} (LLM)_m \times (Holland)_k + \gamma_{0i} + \epsilon_{j,i}.$$
Level 1 model: $$y_{j,i} = \pi_{0i}+\epsilon_{ji},$$
Level 2 model: $$\pi_{0i} = \beta_0 + \beta_1 (LLM)_m+\beta_2 (Holland)_{k}+\beta_{3} (LLM)_m \times (Holland)_k + \gamma_{0i},$$
where $y_{j,i}$ is the $i$-th item's Holland interest score of the $m$-th LLM on the $k$-th dimension of Holland interest category at the $j$-th measure, and $\epsilon_{j,i,j,k}$ is the residual variance of random error. $\gamma_{0i}$ is the residual from predicting item $i$'s mean on each LLM and Holland interest category. 

The four LLMs and the six interest categories were treated as fixed effects on the interest scores of OIP. Then marginal comparisons and pairwise comparisons were performed on each level of the fixed effects in order to get a more detailed look into the relationships of interest scores and the fixed effects, with Tukey adjustments to ameliorate increasing Type I error with multiple comparisons. The intercept of each item was allowed to vary by adding an random intercept parameter $\gamma_{0i}$ into the model to account for the variations among the 20 repeated measures for each item. Since the effect of repeated measures were not the focus of the study, no other level-1 predictors were introduced in the model.

\bibliographystyle{sn-nature}



\section*{Supplementary material}

# The Career Interest of Large Language Models

**Supplementary 1: parameter estimates of hierarchical models**

**Table S1.1**: Mixed model parameter estimates for interest scores by the RIASEC and LLM families, with the Realistic category and gpt-3.5 as the reference groups.

| Fixed Effects | Coefficient | SE | t | df | p |
|---|---|---|---|---|---|
| Model for initial status, | | | | | |
| Mean interest score | 3.530 | 0.122 | 29.041 | 4560 | <.001 |
| **LLM** | | | | | |
| Gemini-pro | -0.210 | 0.180 | -1.168 | 216 | 0.244 |
| ERNIE-3.5 | -0.565 | 0.172 | -3.288 | 216 | 0.001 |
| Spark-1.5 | 0.030 | 0.178 | 0.169 | 216 | 0.866 |
| GPT-3.5 | | | | | |
| **Holland Categories** | | | | | |
| Investigative | 0.025 | 0.172 | 0.145 | 216 | 0.885 |
| Artistic | 0.300 | 0.171 | 1.752 | 216 | 0.081 |
| Social | 0.390 | 0.170 | 2.289 | 216 | 0.023 |
| Enterprising | -0.260 | 0.172 | -1.515 | 216 | 0.131 |
| Conventional | -0.115 | 0.171 | -0.672 | 216 | 0.502 |
| Realistic | | | | | |
| **Random Effects** | | | | | |
| Variance Component | **Coefficient** | **Std** | | | |
| Intercept | .139 | .343 | | | |
| Level-1 error | .169 | .410 | | | |

Note: Likelihood ratio test for the model comparison of mixed model with random effect and without random effect showed that $\chi^2 = 2270.187$, df = 24, p < .0001.

**Table S1.2**: Mixed model parameter estimates for interest scores by Language controlling for the RIASEC, and LLM families with the Realistic category, gpt-3.5 and Chinese as the reference groups.

| Fixed Effects | Coefficient | SE | t | df | p |
| --- | --- | --- | --- | --- | --- |
| Model for initial status, | | | | | |
| Mean interest score | 3.585 | .121 | 29.63 | 9120 | <.001 |
| **LLM** | | | | | |
| Gemini-pro | .010 | .170 | .059 | 432 | .953 |
| ERNIE-3.5 | -.005 | .172 | -.029 | 432 | .977 |
| Spark-1.5 | -.285 | .168 | -1.693 | 432 | .091 |
| GPT-3.5 | | | | | |
| **Holland Categories** | | | | | |
| Investigative | .105 | .170 | .618 | 432 | .537 |
| Artistic | .385 | .171 | 2.258 | 432 | .025 |
| Social | .180 | .171 | 1.054 | 432 | .292 |
| Enterprising | -.140 | .171 | -.821 | 432 | .412 |
| Conventional | -.330 | .173 | -1.911 | 432 | .057 |
| **Language** | | | | | |
| English | -.06 | .169 | -.32 | 432 | .746 |
| **Random Effects** | | | | | |
| Variance Component | **Coefficient** | **Std** | | | |
| Intercept | .132 | .363 | | | |
| Level-1 error | .723 | .850 | | | |

Note: Likelihood ratio test for the model comparison of mixed model with random effect and without random effect showed that χ² = 4834, df = 48, p < .0001.

**Table S1.3**: Mixed model parameter estimates for interest scores by GPT version advances controlling for the RIASEC and language with gpt-3.5-original, Realistic category, and Chinese as the reference groups.

| Fixed Effects | Coefficient | SE | t | df | p |
| --- | --- | --- | --- | --- | --- |
| Model for initial status, | | | | | |
| Mean interest score | 3.275 | .122 | 26.884 | 9120 | <.001 |
| **LLM** | | | | | |
| GPT-0613 | .310 | .175 | 1.773 | 432 | .077 |
| GPT-1106 | .275 | .174 | 1.579 | 432 | .115 |
| GPT-0125 | .580 | .173 | 3.348 | 432 | .001 |
| GPT-origin | | | | | |
| **Holland Categories** | | | | | |
| Investigative | .245 | .173 | 1.417 | 432 | .157 |
| Artistic | .405 | .173 | 2.339 | 432 | .020 |
| Social | .385 | .173 | 2.228 | 432 | .026 |
| Enterprising | -.245 | .172 | -1.427 | 432 | .154 |
| Conventional | -.030 | .173 | -.174 | 432 | .862 |
| **Language** | | | | | |
| English | -.230 | .172 | -1.335 | 432 | .183 |
| **Random Effects** | | | | | |
| Variance Component | Coefficient | Std | | | |
| Intercept | .146 | .382 | | | |
| Level-1 error | .120 | .346 | | | |

Note: Likelihood ratio test for the model comparison of mixed model with random effect and without random effect showed that $\chi^2 = 5918.07$, df = 24, p < .0001.

**Table S1.4**: Mixed model parameter estimates for interest scores by ERNIE version advances controlling for the RIASEC and language with ERNIE-3.5-original, Realistic category, and Chinese as the reference groups.

| Fixed Effects | Coefficient | SE | t | df | p |
|---|---|---|---|---|---|
| Model for initial status, | | | | | |
| Mean interest score | 3.580 | .159 | 22.520 | 6840 | <.001 |
| **LLM** | | | | | |
| ERNIE-1222 | -.355 | .243 | -1.459 | 324 | .146 |
| ERNIE-0205 | -.225 | .224 | -1.003 | 324 | .317 |
| ERNIE-origin | | | | | |
| **Holland Categories** | | | | | |
| Investigative | .375 | .224 | 1.674 | 324 | .095 |
| Artistic | .432 | .224 | 1.939 | 324 | .053 |
| Social | .390 | .225 | 1.737 | 324 | .083 |
| Enterprising | .305 | .224 | 1.362 | 324 | .174 |
| Conventional | | | | | |
| **Language** | | | | | |
| English | -.615 | .225 | -.274 | 324 | .007 |
| **Random Effects** | | | | | |
| Variance Component | Coefficient | Std | | | |
| Intercept | .240 | .490 | | | |
| Level-1 error | 1.983 | 1.408 | | | |

Note: Likelihood ratio test for the model comparison of mixed model with random effect and without random effect showed that $\chi^2 = 5648.32$, df = 18, p < .0001.

**Table 1.5**: Mixed model parameter estimates for interest and competence scores by the RIASEC, LLM families, and language with the Realistic category, gpt-3.5 and Chinese as the reference groups.

| Fixed Effects | Coefficient | SE | t | df | p |
|---|---|---|---|---|---|
| Model for initial status, | | | | | |
| Mean interest score | 4.155 | .154 | 27.028 | 18240 | <.001 |
| **LLM** | | | | | |
| Gemini-pro | -2.095 | .226 | -9.250 | 864 | <.001 |
| GPT-3.5 | -.405 | .219 | -1.853 | 864 | .064 |
| Spark-1.5 | -2.030 | .230 | -8.826 | 864 | <.001 |
| ERNIE-3.5 | | | | | |
| **Holland Categories** | | | | | |
| Realistic | -.020 | .216 | -.093 | 864 | .926 |
| Investigative | -.165 | .216 | -.763 | 864 | .446 |
| Social | -.170 | .215 | -.789 | 864 | .430 |
| Enterprising | -.150 | .215 | -.696 | 864 | .486 |
| Conventional | -.055 | .214 | -.256 | 864 | .798 |
| Artistic | | | | | |
| **Language** | | | | | |
| English | -1.140 | .218 | -5.225 | 864 | <.001 |
| **Competence** | | | | | |
| Interest | -.140 | .215 | -.652 | 864 | .515 |
| **Random Effects** | | | | | |
| Variance Component | **Coefficient** | **Std** | | | |
| Intercept | .224 | .473 | | | |
| Level-1 error | .169 | .411 | | | |

Note: Likelihood ratio test for the model comparison of mixed model with random effect and without random effect showed that $\chi^2$ = 13773.6, df = 96, p < .0001.

**Supplementary 2: full text, prompts and rubrics of OIP and competence**

2.1 Full text of the O*NET Interest Profiler (60 items)

Instruction:

Read each question carefully and decide how you would feel about doing each type of work:

- Strongly Dislike
- Dislike
- Unsure
- Like
- Strongly Like

Try NOT to think about:

- If you have enough education or training to do the work; or
- How much money you would make doing the work.

There are no right or wrong answers! Just think about if you would like or dislike doing the following work.

Examine blood samples using a microscope

Answer:

| English | Chinese |
| --- | --- |
| Build kitchen cabinets | 建造厨柜 |
| Lay brick or tile | 铺砖或瓷砖 |
| Develop a new medicine | 研发新药物 |
| Study ways to reduce water pollution | 研究减少水污染的方法 |
| Write books or plays | 撰写书籍或剧本 |
| Play a musical instrument | 演奏乐器 |
| Teach an individual an exercise routine | 教授个人运动计划 |
| Help people with personal or emotional problems | 帮助处理个人或情绪问题 |
| Buy and sell stocks and bonds | 买卖股票和债券 |
| Manage a retail store | 管理零售店 |
| Develop a spreadsheet using computer software | 使用计算机软件开发电子表格 |
| Proofread records or forms | 校对记录或表格 |
| Repair household appliances | 修理家用电器 |
| Raise fish in a fish hatchery | 在鱼苗孵化场养鱼 |
| Conduct chemical experiments | 进行化学实验 |
| Study the movement of planets | 研究行星运动 |
| Compose or arrange music | 创作或编曲音乐 |
| Draw pictures | 绘画 |
| Give career guidance to people | 为人们提供职业指导 |
| Perform rehabilitation therapy | 进行康复疗法 |
| Operate a beauty salon or barber shop | 经营美容沙龙或理发店 |
| Manage a department within a large company | 在大公司内管理一个部门 |
| Install software across computers on a large network | 在大型网络上安装软件 |
| Operate a calculator | 操作计算器 |
| Assemble electronic parts | 组装电子零件 |
| Drive a truck to deliver packages to offices and homes | 驾驶卡车向办公室和家庭递送包裹 |
| Examine blood samples using a microscope | 使用显微镜检查血液样本 |
| Investigate the cause of a fire | 调查火灾原因 |
| Create special effects for movies | 为电影制作特效 |
| Paint sets for plays | 为戏剧绘制舞台布景 |

| | |
|---|---|
| Do volunteer work at a non-profit organization | 在非营利组织做义工 |
| Teach children how to play sports | 教导儿童如何进行体育运动 |
| Start your own business | 创办自己的业务 |
| Negotiate business contracts | 协商商业合同 |
| Keep shipping and receiving records | 记录发货和收货记录 |
| Calculate the wages of employees | 计算员工工资 |
| Test the quality of parts before shipment | 在发货前测试零件的质量 |
| Repair and install locks | 维修和安装锁具 |
| Develop a way to better predict the weather | 开发更好的天气预测方法 |
| Work in a biology lab | 在生物实验室工作 |
| Write scripts for movies or television shows | 为电影或电视节目编写剧本 |
| Perform jazz or tap dance | 表演爵士舞或踢踏舞 |
| Teach sign language to people who are deaf or hard of hearing | 教授手语给听力障碍或重听人士 |
| Help conduct a group therapy session | 协助进行团体治疗会议 |
| Represent a client in a lawsuit | 代表客户处理诉讼案件 |
| Market a new line of clothing | 推广新系列服装 |
| Inventory supplies using a hand-held computer | 使用手持电脑盘点供应品 |
| Record rent payments | 记录租金支付情况 |
| Set up and operate machines to make products | 设置和操作机器进行产品制造 |
| Put out forest fires | 扑灭森林火灾 |
| Invent a replacement for sugar | 发明替代品取代糖 |
| Do laboratory tests to identify diseases | 进行实验室检测以识别疾病 |
| Sing in a band | 在乐队中演唱 |
| Edit movies | 编辑电影 |
| Take care of children at a day-care center | 在托儿所照顾儿童 |
| Teach a high-school class | 教授高中课程 |
| Sell merchandise at a department store | 在百货商店销售商品 |
| Manage a clothing store | 管理服装店 |
| Keep inventory records | 保持库存记录 |
| Stamp, sort, and distribute mail for an organization | 盖章、分类和分发组织的邮件 |

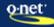

Figure S2.2: prompts for OIP for LLMs

2.3 Rubrics for using the OIP work tasks to evaluate LLM's competence on each task

Large Language Model Competency Assessment Questionnaire

Large language models, as AI assistants, are changing our ways of living and working. They possess powerful natural language processing capabilities, allowing them to understand and generate human language, thus providing help and support in various fields. Large language models have already shown potential as human work assistants.

The following 60 questions represent tasks from six different professional fields. Please read and consider these tasks carefully, then rate the competency of large language models in completing these tasks as work assistants based on your experience.

Competency reflects the knowledge, skills, and abilities needed to complete a task adequately. When rating a task, consider the following three aspects of the large language model as an AI assistant:

1. Does it have the necessary factual and procedural knowledge for the task?
2. Does it have the capability to execute the task accurately?
3. Does it have the cognitive, sensory, and physical abilities required to perform the task?

Please use the following five levels to rate the competency of the large language model as an assistant for each task:

- **Fully Competent**: The large language model can help you complete the task comprehensively and improve the quality or significantly reduce the time required (at least by half).
- **Partially Competent**: The large language model can help you complete parts of the task well, improve the quality, or reduce the time required.
- **Unsure**: The large language model may or may not help, somewhere between competent and incompetent.

- **Not Very Competent**: The large language model can barely help you complete the task, cannot improve the quality or reduce the time required.
- **Not Competent at All**: The large language model cannot help you complete the task at all, and may even lower the quality and increase the time required.

Please mark the appropriate level.



\end{document}